\documentclass[10pt,twocolumn,letterpaper]{article}

\usepackage[pagenumbers]{iccv}      
\usepackage{multirow}
\usepackage{indentfirst}
\newenvironment{sizeddisplay}[1]
 {\par\nopagebreak#1\noindent\ignorespaces}
 {\nopagebreak\ignorespacesafterend}
\definecolor{iccvblue}{rgb}{0.21,0.49,0.74}
\usepackage[pagebackref,breaklinks,colorlinks,allcolors=iccvblue]{hyperref}
\usepackage{adjustbox}
\def\paperID{14789} 
\def\confName{ICCV}
\def\confYear{2025}

\title{Enhancing Variational Autoencoders with Smooth Robust Latent Encoding}

\author{
  Hyomin Lee\textsuperscript{1}\thanks{Equal contribution} \thanks{Work done during an internship at KAIST}\quad 
  Minseon Kim\textsuperscript{2}\footnotemark[1] \quad
  Sangwon Jang\textsuperscript{3}\quad
  Jongheon Jeong\textsuperscript{1} \quad
  Sung Ju Hwang\textsuperscript{3,4} \\
  \textsuperscript{1}Korea University, \textsuperscript{2}Microsoft, \textsuperscript{3}KAIST, \textsuperscript{4}DeepAuto.ai \\
  \normalsize
  \texttt{\{lhm1024, jonghj\}@korea.ac.kr, minseonkim@microsoft.com}, \\ 
  \normalsize
  \texttt{\{sangwon.jang, sungju.hwang\}@kaist.ac.kr}
}

\begin{document}
\maketitle
\begin{abstract}
Variational Autoencoders (VAEs) have played a key role in scaling up diffusion-based generative models, as in Stable Diffusion, yet questions regarding their robustness remain largely underexplored.
Although adversarial training has been an established technique for enhancing robustness in predictive models, it has been overlooked for generative models due to concerns about potential fidelity degradation by the nature of trade-offs between performance and robustness. In this work, we challenge this presumption, introducing Smooth Robust Latent VAE (SRL-VAE), a novel adversarial training framework that boosts both generation quality and robustness. In contrast to conventional adversarial training, which focuses on robustness only, our approach smooths the latent space via adversarial perturbations, promoting more generalizable representations while regularizing with originality representation to sustain original fidelity. Applied as a post-training step on pre-trained VAEs, SRL-VAE improves image robustness and fidelity with minimal computational overhead. Experiments show that SRL-VAE improves both generation quality, in image reconstruction and text-guided image editing, and robustness, against Nightshade attacks and image editing attacks. These results establish a new paradigm, showing that adversarial training, once thought to be detrimental to generative models, can instead enhance both fidelity and robustness.
\end{abstract}

\section{Introduction}
Variational Autoencoders (VAEs)~\citep{kingma2019VAE} have been employed as a compressor in the success of latent generative models~\citep{rombach2022high, peebles2023scalable, ma2024sit, li2024autoregressive}, which have demonstrated surprising capabilities in generating high-quality images. The VAEs compress high-dimensional images into a latent space that retains semantic and structural information, which is continuous~\citep{rombach2022high} or discrete~\citep{esser2021taming} space for high-quality generative modeling. Despite their effectiveness as a compressor in generative models, prior work has largely overlooked the representational role of VAEs, primarily focusing on generative aspects to improve performance by proposing architectures~\citep{peebles2023scalable}, training objectives~\citep{xiang2023denoising, ma2024sit}, or regularization~\citep{yu2024representation}. However, obtaining an effective compressor is one of the key components to achieving higher-quality generations while also ensuring robustness with efficient computational costs.

\begin{figure}[t]
    \centering
    \begin{subfigure}{\linewidth}
        \centering
        \includegraphics[width=0.8\linewidth]{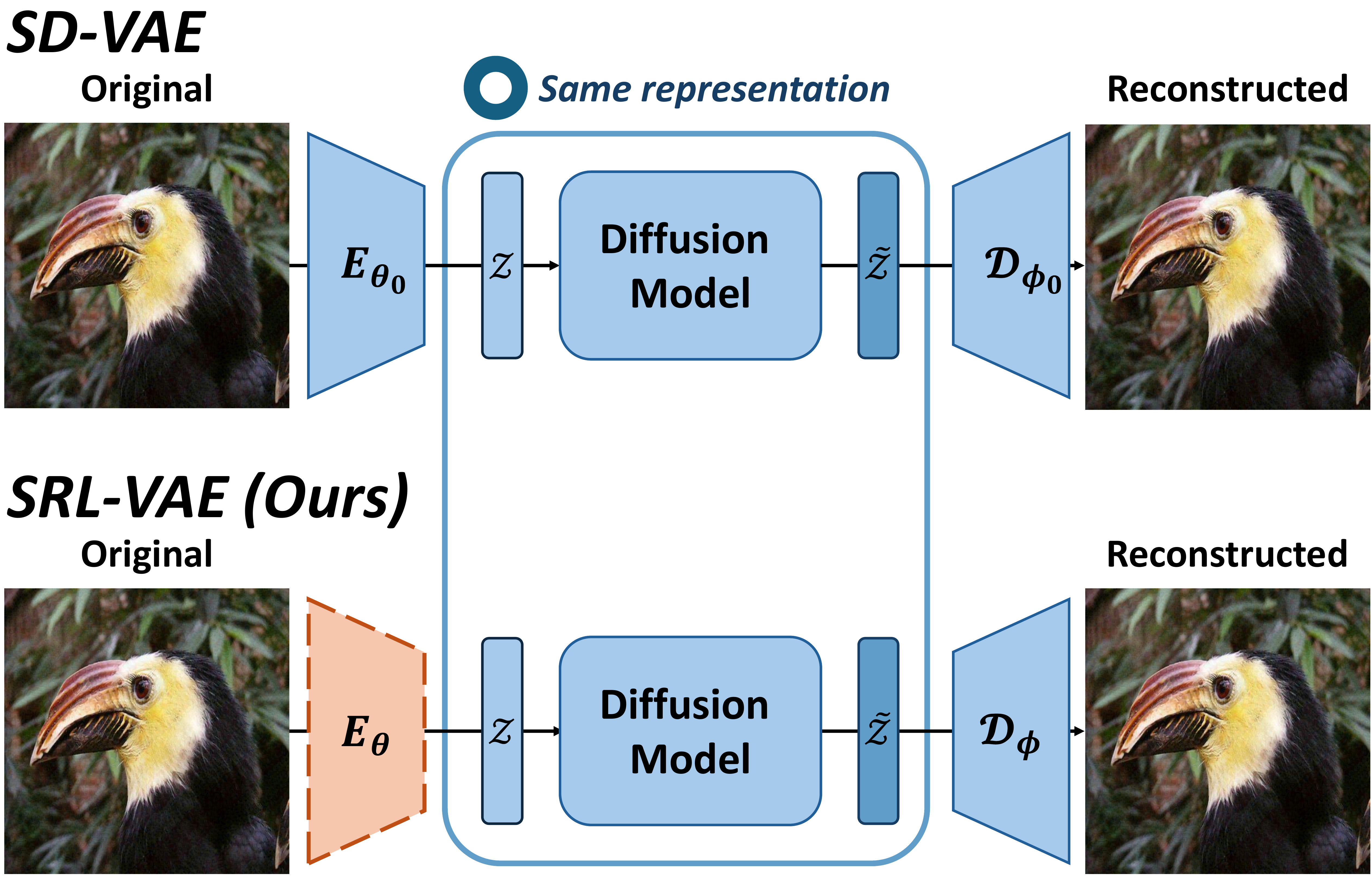}
        \caption{Diffusion-based generative process of clean image}
        \vspace{0.07in}
        \label{figure:ours1}
    \end{subfigure}
    \begin{subfigure}{\linewidth}
        \centering
        \includegraphics[width=0.8\linewidth]{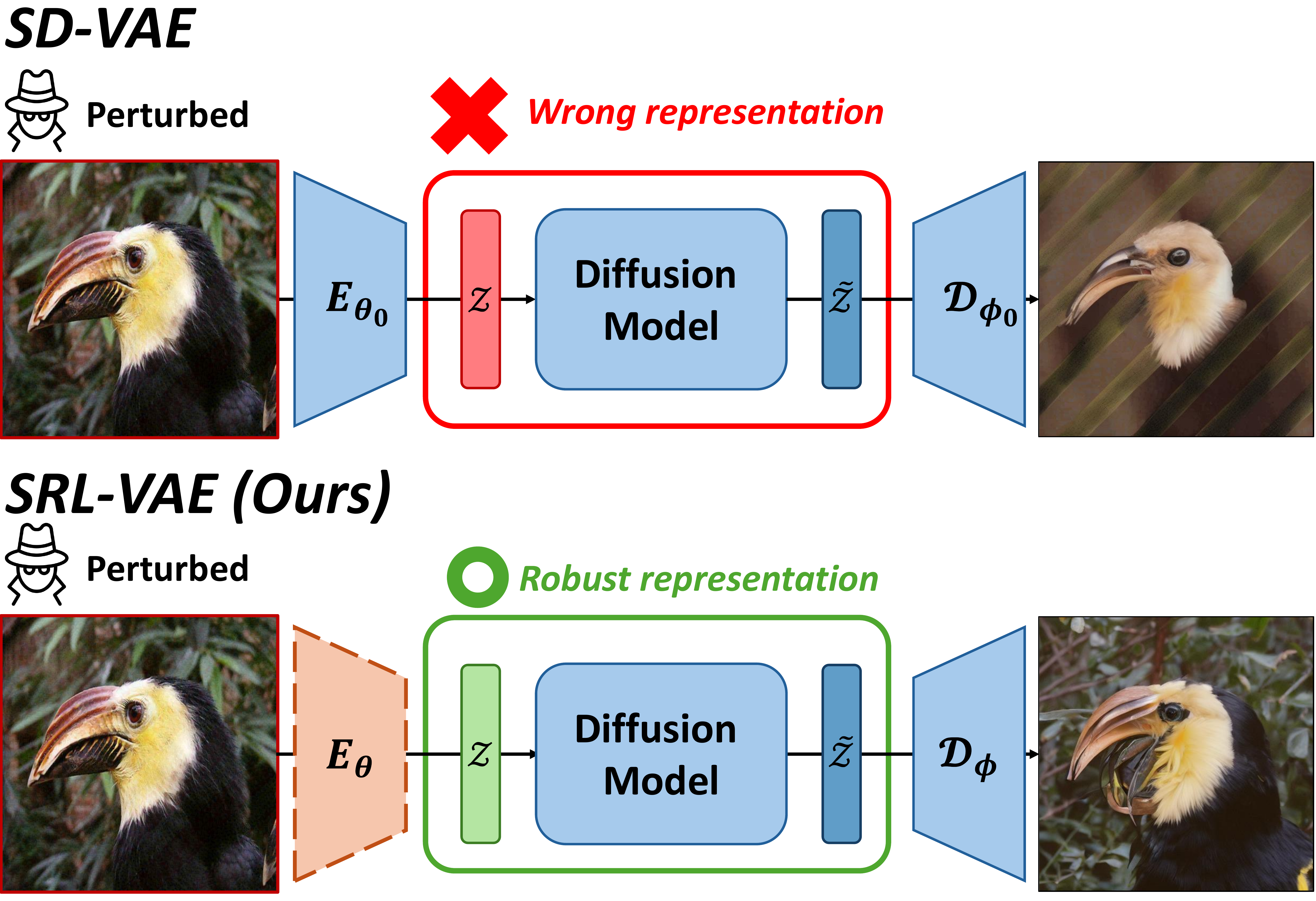}
        \caption{Diffusion-based generative process of perturbed image}
        \label{figure:ours2}
    \end{subfigure}
    \vspace{-0.2in}
    \caption{\textbf{Concept figure of SRL-VAE.} Compared to SD-VAE, SRL-VAE maintains similar representations for clean examples while achieving robust representation against perturbed examples.}
    \vspace{-0.15in}
    \label{figure:ours}
\end{figure}

To obtain effective VAEs for both higher fidelity and better robustness, representation space of VAEs needs to be robust and capable of capturing better structural latent. We were inspired by adversarial training~\citep{madry2017pgd, zhang2019trades} to build robust representations. Adversarial training~\citep{madry2017pgd, zhang2019trades} has initially been recognized as an effective method for improving adversarial robustness in predictive models against adversarial attacks, particularly in classification tasks. In fact, adversarial training builds robust and smooth representations~\citep{visualloss} so that it could have better generalization~\citep{wu2020awp, kim2020rocl} from leveraging a min-max formulation. However, adversarial training in generative modeling remains largely unexplored, partly because generative models have traditionally prioritized fidelity and diversity over robustness. Moreover, adversarial training is challenging to optimize and often leads to large performance degradation where trade-offs are clear~\citep{zhang2019trades, carmon2019RST}. As a result, the adoption of adversarial training in VAEs and similar generative models has been limited, as it is frequently perceived as a significant trade-off.

\begin{figure}[t]
    \centering
        \centering
        \includegraphics[width=0.8\linewidth]{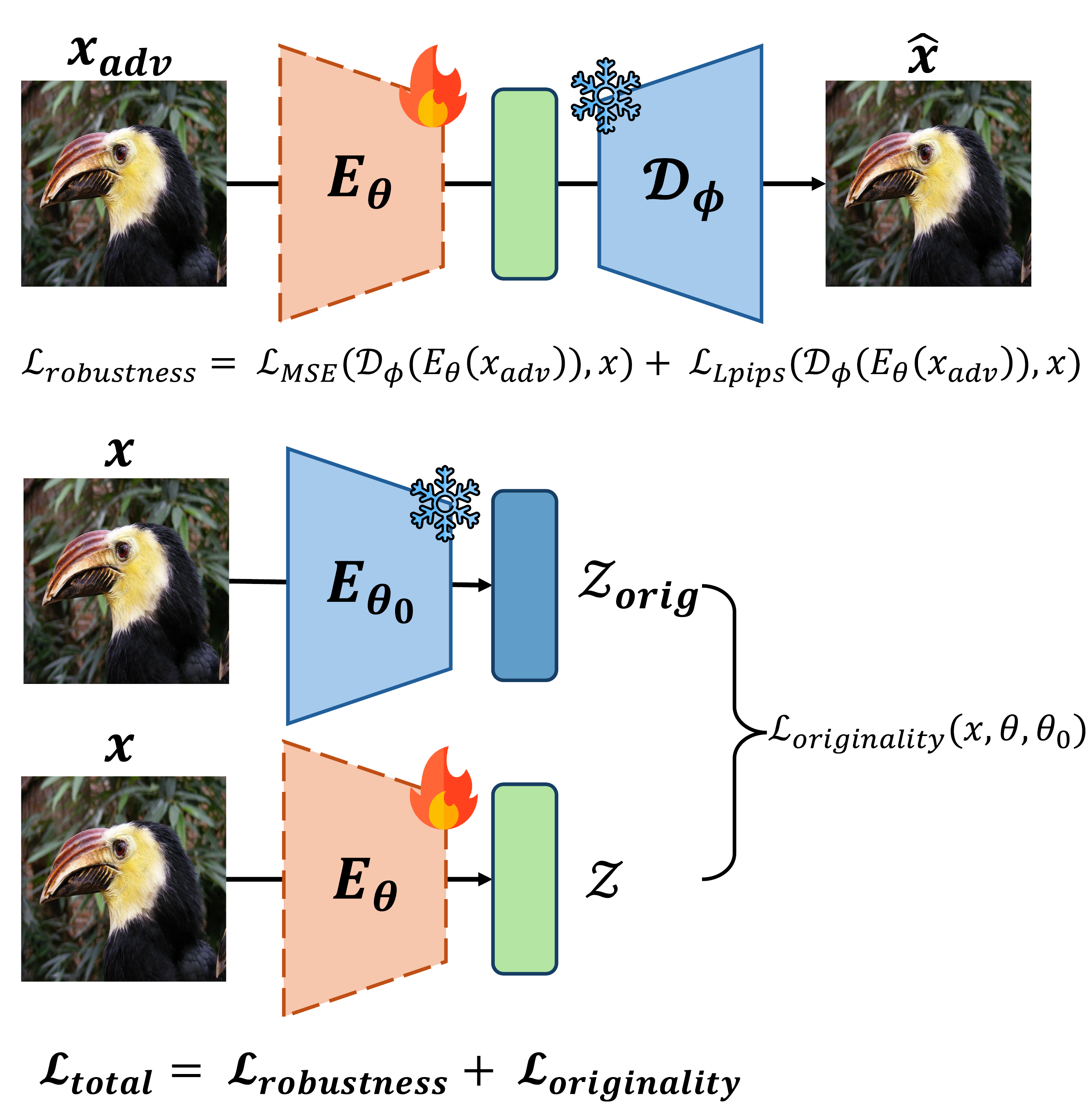}
    \vspace{-0.1in}
    \caption{\textbf{Training objective of Smooth Robust Latent Variational Autoencoders (ours).} A novel adversarial training approach in the latent space of VAE with originality regularization.}
    \label{figure:training}
    \vspace{-0.1in}
\end{figure}

In this work, we introduce a novel adversarial training framework for VAEs that enhances both generation quality and robustness by constructing smooth latent space. Unlike conventional adversarial training, which primarily targets predictive robustness, our approach leverages adversarial perturbations to smooth the latent space and promote more generalizable representations of VAEs (Figure~\ref{figure:ours}). Our approach consists of two key steps: (1) maximizing the VAE loss to introduce adversarial perturbations in the latent space, exposing the model to challenging variations, and (2) minimizing both the VAE loss and an originality loss to preserve the original representation structure while building smooth latent space, ensuring a stable training and robustness as shown in Figure~\ref{figure:training}. Furthermore, our approach is applied as a post-training step on pre-trained VAEs, requiring only a small amount of additional computational resources, making it an efficient and practical solution for improving generative models. 

By bridging adversarial training and generative modeling, our work introduces a new perspective on the importance of obtaining robust, high-quality representations of VAEs in generative models. Our method enables VAEs to generate outputs of comparable or, in some cases, even higher quality (Figure~\ref{figure:ours1}) while being extremely effective against various types of adversarial attacks during the generation process (Figure~\ref{figure:ours2}). This demonstrates that adversarial training is not merely a defensive mechanism that sacrifices performance but rather a powerful strategy for enhancing generative models with robustness, paving the way for future advancements in robust generative learning.

The main contributions can be summarized as follows:
\begin{itemize}
    \vspace{0.05in}
    \item Unlike prior adversarial training methods, which predominantly focus on classification models, we introduce the first adversarial training approach tailored for VAEs, demonstrating its ability to improve both generation quality and robustness simultaneously.
    \setlength{\itemsep}{4pt}
    \item We show that \textbf{adversarial training}, when combined with an originality loss, \textbf{fosters a smoother latent space, leading to more stable and generalizable representations,} which enhance image fidelity and deliver surprisingly strong performance against various types of attacks.
    \setlength{\itemsep}{4pt}
    \item Through extensive experiments, we demonstrate significant improvements in both fidelity and robustness across multiple tasks. Specifically, we evaluate image quality through image reconstruction and generation, and assess robustness by evaluation against adversarial attacks on text-guided image editing and adversarial poisoning attacks, establishing the effectiveness of our approach in enhancing both generative performance and robustness.
\end{itemize}
\section{Related Works}

\paragraph{Latent Generative Models}
Variational Autoencoders (VAEs)~\citep{kingma2019VAE} are generative models that learn compact latent representations by regularizing the latent distribution through a Kullback–Leibler (KL) divergence term. While VAEs enable smooth interpolation and efficient sampling, they often produce blurry outputs due to limitations in the latent space. To address this, Vector Quantized VAE (VQ-VAE)~\citep{van2017neural} introduces a discrete codebook of embeddings, preventing latent space collapse and improving reconstruction quality. VQ-GAN~\citep{esser2021taming} further enhances this framework with adversarial training, guiding the decoder towards sharper and more realistic outputs. Building on these approaches, Latent Diffusion Models (LDMs)~\citep{rombach2022high} apply diffusion processes in a compressed latent space obtained from a pretrained autoencoder, significantly reducing computational costs while preserving high fidelity. Cross-attention mechanisms enable flexible conditional generation from various inputs, as demonstrated in Stable Diffusion~\citep{rombach2022high}. Recent work also explores enhancing the interaction between autoencoders and diffusion models, addressing spectral properties of latent spaces ~\citep{skorokhodov2025improving} and introducing equivariance regularization for better generative performance ~\citep{kouzelis2025eq}. However, prior works have largely overlooked the quality of latent representations in VAEs in terms of fidelity and robustness. We address this by applying adversarial training to enhance the latent space, improving both generation quality and robustness.
\vspace{-0.12in}
\paragraph{Adversarial Training}
\citet{szegedy2013intriguing} first revealed the vulnerability of deep neural networks (DNNs) to adversarial attacks, showing that imperceptible perturbations could mislead models. \citet{goodfellow2014fgsm} introduced adversarial training with the Fast Gradient Sign Method (FGSM), demonstrating that training on adversarial and clean samples improves robustness. \citet{madry2017pgd} extended this with Projected Gradient Descent (PGD) adversarial training, formulating it as a minimax optimization problem. TRADES~\citep{zhang2019trades} further refined robustness by enforcing consistency between clean and adversarial samples via Kullback-Leibler divergence minimization. Recent works leveraged unlabeled data~\citep{carmon2019RST} or generative models~\citep{gowal2021improving} to enhance adversarial robustness by exposing diverse distribution. Beyond supervised settings, adversarial self-supervised learning (SSL) emerged as an alternative perspective to obtain robust representation~\citep{jiang2020ACL, kim2023effective}, using contrastive learning or self-supervised learning by introducing adversarial examples that maximize given losses without any class information. However, all these methods focus on building a robust representation for predictive models to have robust decision boundaries, which did not consider generative models. Unlike prior works, our approach suggests adversarial training for VAEs, demonstrating that it can enhance both generation quality and robustness.

\section{Variation Autoencoders with Smooth Robust Latent Encoding}
In this section, we first revisit the preliminary of Variational Autoencoder (VAE) and adversarial training (AT) in section~\ref{sec:prelim}. Then, we propose our smooth robust latent VAE approach with theoretical motivation in Section~\ref{sec:ours}.

\subsection{Preliminary~\label{sec:prelim}}
\paragraph{Variational autoencoder (VAE)}
A Variational autoencoder (VAE) is a latent space compressor that encodes high-dimensional image data into a lower-dimensional latent space. Given an input image $x$, the encoder $E_\theta(x)$ compresses the image to a latent variable $z$, and the decoder $D_\phi(z)$ reconstructs $x$ as $\hat{x} = D_\phi(z)$. We employ a VAE~\citep{rombach2022high} that is optimized primarily for high-fidelity reconstruction. The training objective is defined as follows:
\begin{equation}
    \mathcal{L}_\texttt{VAE}(x) = \mathcal{L}_\texttt{rec}(x, \hat{x}) + \mathcal{L}_\texttt{gan}(\hat{x}) + \mathcal{L}_\texttt{reg}(x),
\end{equation}
where $\mathcal{L}_\texttt{rec}$ combines pixel-wise loss ($L_1$ distance loss or $L_2$ distance loss) and perceptual loss (LPIPS loss). LPIPS loss is a similarity loss of learned perceptual image patches that calculates the similarity distance based on features extracted from a pre-trained VGG model~\citep{simonyan2014very}. The $\mathcal{L}_\texttt{gan}$ is an adversarial loss that encourages the generation of more realistic outputs by leveraging a discriminator network. Lastly, $\mathcal{L}_\texttt{reg}$ is a regularization term which is following KL-divergence:
\begin{equation}
    \mathcal{L}_\texttt{reg}(x) = \mathcal{L}_\texttt{KL}(q_\theta(z|x) || p(z)),
\end{equation}
where $p(z)$ is the prior distribution, and set to a standard normal distribution $\mathcal{N}(0, I)$. KL regularization in recent latent diffusion models ensures smooth sampling and interpolation by encouraging a well-structured latent space.

\vspace{-0.12in}
\paragraph{Adversarial training}
Adversarial training is a technique for obtaining robust models against adversarial attacks by solving a min-max optimization problem. First, we define an adversarial perturbation, $\delta$, which is applied to an input $x$. Then, the min-max optimization problem is formulated with 1) generating the perturbation $\delta$ by maximizing the model's given loss $\mathcal{L}$, while 2) simultaneously minimizing the training loss under this perturbation. Several approaches~\citep{zhang2019trades, wang2019mart} exist for generating effective adversarial perturbation in the min-max optimization problem. Here, we employ projected gradient descent attacks~\citep{madry2017pgd} that maximize the training loss, as follows.
\begin{sizeddisplay}{\small}
\begin{equation}
\begin{aligned}
     \delta^{t+1} = \Pi_{B(0,\epsilon)} \Big(\delta^t + \alpha \texttt{sign}\Big(\nabla_{\delta^t} \mathcal{L}\Big)\Big),
      \label{equation:pgd_attacks}
\end{aligned}
\end{equation}
\end{sizeddisplay}
where $B(0,\epsilon)$ is the $\ell_{\infty}$ norm-ball of radius $\epsilon$, $\Pi$ is the projection function to the norm-ball, $\alpha$ is the step size of the attacks, and $\texttt{sign}(\cdot)$ is the sign of the vector. Also, $\delta$ represents the perturbations accumulated by $\alpha \texttt{sign}(\cdot)$ over multiple iterations $t$. Then, minimization is defined as follows,
\begin{equation}
    \min_{\omega} \mathbb{E}_{x\sim \mathcal{D}} \left[\mathcal{L}(x + \delta^t) \right],
\end{equation}
where $\omega$ is a parameter of model $f$, $x$ is training samples from dataset $\mathcal{D}$ and $\mathcal{L}$ is training objectives which employ perturbed samples $x+\delta$.

\subsection{Smooth Robust Latent VAE~\label{sec:ours}}
\vspace{-0.03in}
\begin{figure}[t]
    \centering
        \centering
        \includegraphics[width=\linewidth]{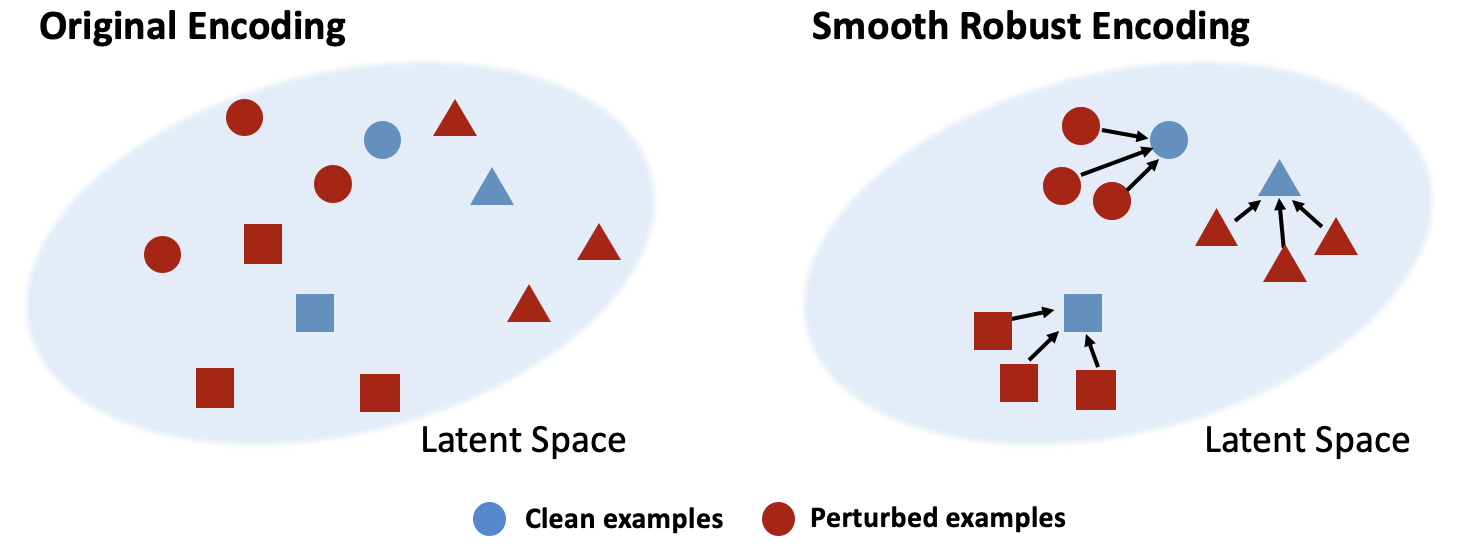}
    \vspace{-0.2in}
    \caption{\textbf{Concept of smooth latent space.} A smooth latent space ensures that perturbed examples are mapped closely to their original counterparts, enabling the VAE to extract robust features.}
    \label{figure:robust_loss}
    \vspace{-0.1in}
\end{figure}

\paragraph{Theoretical motivation}
Adversarial training has been widely used in predictive models to improve robustness against adversarial attacks, yet its application in generative models remains under-explored. In predictive tasks, robustness is achieved by ensuring that perturbed inputs $x + \delta$ produce outputs similar to those of the original inputs $x$. This is typically enforced by a Lipschitz constraint, which guarantees that small changes within an $\epsilon$-norm ball result in only minor variations in the output, thereby creating a smooth representation space. Motivated by this, we propose that applying adversarial training to VAEs can similarly promote a smooth and well-structured latent space, leading to improved generation quality and robustness (Figure~\ref{figure:robust_loss}).

Furthermore, in latent-based generative models, the encoder acts as an information bottleneck, compressing high-dimensional inputs into latent codes $z = E_\theta(x)$. Motivated by the Information Bottleneck (IB) principle, we believe that an optimal latent representation should capture only the essential features for accurate reconstruction while discarding noisy features and clearly distinguishing different inputs. In VAEs, this results in a latent space that is both expressive and compact. By applying adversarial training, our approach forces the encoder to extract only the crucial features for high-quality reconstruction and to maintain clear separations among examples. Specifically, adversarial training encourages each input's latent representation to be confined within a secure $\epsilon$-ball, creating a large margin between different examples. This leads to a tighter information bottleneck and a more structured latent space, ultimately enhancing both image fidelity and generalization.


\vspace{-0.12in}
\paragraph{Smooth Robust Latent VAE}
We propose \textit{Smooth Robust Latent VAE} (SRL-VAE), which enhances latent representation quality by applying adversarial training to the encoder. Additionally, to ensure compatibility with pre-trained diffusion models such as the UNet in latent diffusion models (LDMs), our approach emphasizes preserving the original latent structure while refining it for improved performance.

We formulate a min-max optimization framework for VAEs. In particular, to generate adversarial perturbations, we define the maximization step using a projected gradient descent (PGD) formulation as follows:
\begin{sizeddisplay}{\small}
\begin{equation}
\begin{aligned}
     \delta^{t+1} = \Pi_{B(0,\epsilon)} \Big(\delta^t + \alpha \texttt{sign}\Big(\nabla_{\delta^t} (\mathcal{L}_\texttt{MSE}(D_\phi(E_\theta(x+\delta^t)), x) \\
    + \lambda \cdot \mathcal{L}_\texttt{LPIPS}(D_\phi(E_\theta(x+\delta^t)), x))\Big)\Big),
      \label{equation:attacks}
\end{aligned}
\end{equation}
\end{sizeddisplay}
where $\mathcal{L}_\texttt{MSE}$, and $\mathcal{L}_\texttt{LPIPS}$ is $L_2$ distance loss and LPIPS perceptual loss between adversarial examples and original examples, respectively. 
    
Then, the minimization step is formulated to ensure that the outputs from the perturbed examples are similar to the original examples, while preserving the original latent space distribution. This is expressed as follows:
\begin{equation}
\begin{aligned}
\mathcal{L}_{\texttt{total}} = \alpha \mathcal{L}_{\texttt{orig}}(x, \theta, \theta_0)
    +\mathcal{L}_\texttt{MSE}(D_\phi(E_\theta(x_{\texttt{adv}})), x) \\
        + \lambda \mathcal{L}_\texttt{LPIPS}(D_\phi(E_\theta(x_{\texttt{adv}})), x)
    \end{aligned}
\label{equation:total}
\end{equation}
where $x_{\texttt{adv}} = x + \delta$ from equation~\ref{equation:attacks}, and $\alpha$, $\lambda$ control the balance between latent consistency and reconstruction quality. 

The originality loss $\mathcal{L}_{\texttt{orig}}$ acts as regularization to preserve the latent distribution of clean inputs by minimizing the difference from the pre-trained encoder:
\begin{equation}
\mathcal{L}_{\texttt{orig}}(x, \theta, \theta_0) = \| \mu - \mu_{\texttt{orig}} \|_2^2 + \| \log \sigma^2 - \log \sigma^2_{\texttt{orig}} \|_2^2,
\end{equation}
where $\mu$ and $\sigma^2$ represent the mean and variance of the latent distribution produced by the current encoder parameterized by $\theta$, and $\mu_\texttt{orig}$ and $\sigma_\texttt{orig}^2$ represent the mean and variance produced by the pre-trained encoder parameterized by $\theta_0$, respectively, when given input $x$. By minimizing this objective, the encoder learns robust latent representations that maintain the original latent distribution, ensuring compatibility with downstream components and enabling stable, high-fidelity generation with enhanced robustness.

\section{Experiment}
In this section, we first describe our experimental setup, including datasets, training details, and evaluation details in Section~\ref{sec:setup}. We then present the image quality performance of our SRL-VAE in Section~\ref{sec:image_quality}, demonstrating both its reconstruction quality and diffusion generation quality. In Section~\ref{sec:robustness}, we evaluate the robustness of our latent space against various types of perturbations and different attacks in diffusion models. Lastly, we conduct ablation studies and analyze the latent space of SRL-VAE in Section~\ref{sec:analysis}.

\subsection{Setup~\label{sec:setup}}
\paragraph{Training details}
We further fine-tune a pre-trained Stable Diffusion
Variational Autoencoder (SD-VAE) on a subset of 100K images from the LAION-Aesthetic dataset~\citep{schuhmann2022laion}, resized to 256$\times$256 resolution. During fine-tuning, only the encoder is updated while keeping the decoder frozen to maintain compatibility with the pre-trained diffusion model. The model is optimized with a batch size of 20 for a total of 5K steps. For adversarial training, we use the Projected Gradient Descent (PGD) attack under an $\ell_\infty$ perturbation bound of $\epsilon = 8/255$ with 10 iterations per attack and a step size of 0.02. We apply the originality loss with a weight of $\alpha = 0.01$, selected through hyperparameter tuning.

\paragraph{Evaluation}
To assess the image quality of SRL-VAE, we evaluate both reconstruction quality and generation quality. For reconstruction quality, we use the MS-COCO~\citep{lin2014coco} validation set (5,000 images) and the ImageNet~\citep{imagenet} validation set (50,000 images), with all images resized to 256×256. We measure Fréchet Inception Distance (FID)~\citep{heusel2017fid}, Peak Signal-to-Noise Ratio (PSNR), Structural Similarity Index (SSIM)~\citep{Wang2004ssim}, and Learned Perceptual Image Patch Similarity (LPIPS)~\citep{zhang2018lpips} to quantify the model’s ability to accurately reconstruct images. For generation quality, we compare SRL-VAE with SD-VAE within the Diffusion Transformer~\citep{peebles2023scalable} (DiT-B/2) framework. We train DiT models on ImageNet-1000k (1,280K images) for 10 epochs (50K steps) with a batch size of 256 and evaluate their performance using Inception Score (IS)~\citep{salimans2016inception} and FID.

To validate the robustness of SRL-VAE against adversarial perturbations, we conducted two experiments. First, we evaluated the model’s resilience to adversarial attacks that target the training process to maliciously manipulate the diffusion model~\citep{shan2024nightshade,lu2024disguised}. Specifically, we tested Nightshade~\citep{shan2024nightshade}, which poisons a concept $C$ so that it generates images resembling a destination concept $A$. To determine attack success, we measured the CLIP~\citep{radford2021clip} similarity between the generated image and its generating prompt $C$, assessing how far the image deviates from the intended concept of $C$. If the CLIP score was lower than the threshold $\tau=0.25$, we considered the attack successful. Second, we assessed the robustness of SRL-VAE against defensive perturbations from PhotoGuard~\citep{salman2023raising}, MIST~\citep{liang2023mist} and Glaze~\citep{shan2023glaze}, which prevent unauthorized image edits. In a realistic image-to-image editing scenario, we compared SRL-VAE and SD-VAE, measuring FID and CLIP similarity with the original generation results to evaluate robustness.

\subsection{Smooth Robust Latent VAE~\label{sec:image_quality}}

\paragraph{Image reconstruction quality}
\begin{table}[t!]
    \centering
    \begin{adjustbox}{width=\linewidth}
    \begin{tabular}{llcccc}
    \toprule
    \textbf{Dataset} & \textbf{VAE} & \textbf{PSNR}~$\uparrow$ & \textbf{SSIM}~$\uparrow$ & \textbf{LPIPS}~$\downarrow$ & \textbf{rFID}~$\downarrow$ \\
    \midrule
    \multirow{2}{*}{COCO} 
        & SD-VAE       & 23.68 & 0.74 & \textbf{0.14} & 8.79 \\
        & Ours       & \textbf{24.46} & \textbf{0.76} & 0.15 & \textbf{7.92} \\
    \midrule
    \multirow{2}{*}{ImageNet} 
        & SD-VAE       & 23.53 & 0.73 & \textbf{0.15} & 3.71 \\
        & Ours       & \textbf{24.48} & \textbf{0.74} & 0.16 & \textbf{3.09} \\
    \bottomrule
    \end{tabular}
    \end{adjustbox}
    \caption{\textbf{Quantitative evaluation of reconstruction quality on the MS-COCO and ImageNet validation sets.} SRL-VAE consistently outperforms the baseline VAE across PSNR, SSIM, and FID metrics, indicating superior fidelity.}
    \label{tab:reconstruction_coco_imagenet}
\end{table}

To evaluate the effectiveness of our proposed SRL-VAE, we measure its performance on image reconstruction tasks compared to the baseline VAE in both COCO dataset~\citep{lin2014coco} and ImageNet dataset (Table~\ref{tab:reconstruction_coco_imagenet}). Our SRL-VAE consistently achieves superior performance across most of the metrics compared to the original VAE. The higher PSNR and SSIM scores indicate that images reconstructed by SRL-VAE preserve more structural and visual details, whereas the lower FID scores demonstrate enhanced perceptual realism of the generated images. These results collectively confirm that adversarial training within our SRL-VAE significantly enhances image fidelity, underscoring the improved quality and robustness of the learned latent representations.

\paragraph{Diffusion generation quality}
We further evaluate the diffusion generation capabilities with our SRL-VAE within the Diffusion Transformer (DiT) framework~\citep{peebles2023scalable}. Our primary objective is to confirm that our adversarial training approach does not compromise generation performance, particularly in the diffusion process. The original DiT model with a standard SD-VAE achieves an IS of 12.49 and a FID of 91.54. In comparison, DiT with our SRL-VAE achieves slightly improved performance, with an IS of 12.87 and an FID of 91.27, demonstrating that our method does not degrade diffusion generation quality. This result highlights that integrating SRL-VAE into the diffusion generation process is both seamless and adaptable, preserving image quality while simultaneously providing additional benefits such as improved latent space representation and enhanced robustness against perturbations, as discussed in Section~\ref{sec:robustness}.

\subsection{Robustness on Perturbations}~\label{sec:robustness}
In this section, we assess the robustness of SRL-VAE by integrating our encoder into existing frameworks and evaluating against two distinct categories of perturbation-based approaches. First, we test against the Nightshade attack~\citep{shan2024nightshade}, a malicious adversarial perturbation-based data poisoning technique designed to disrupt specific outputs of diffusion models by injecting a few poison samples into training data. Second, we evaluate robustness by measuring the neutralization scale against defensive perturbation methods such as PhotoGuard~\citep{salman2023raising}, Glaze~\citep{shan2023glaze}, and Mist~\citep{liang2023mist}, which are initially designed to protect intellectual property by adding imperceptible perturbations. Our experiments demonstrate that our SRL-VAE effectively neutralizes both types of perturbation-based approaches.
\vspace{-0.12in}
\paragraph{Robustness in Nightshade malicious attack}
\begin{figure*}[t]
\centering
\begin{minipage}{0.22\linewidth}
    \centering
    \includegraphics[width=1.0\linewidth]{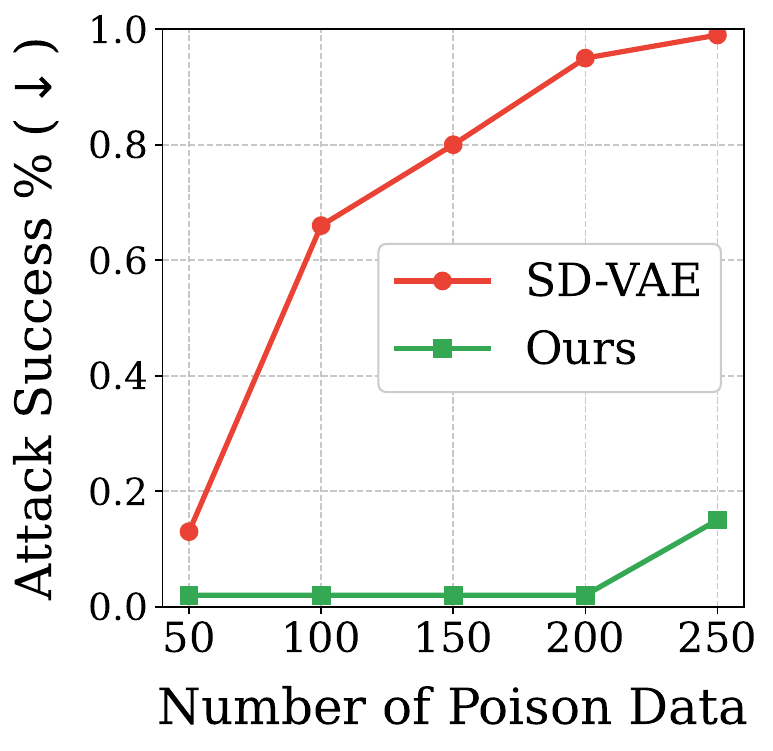}
    \vspace{-0.28in}
    \centering
    \caption*{\textbf{(a)} Attack success rate depending on \textbf{number of poisoned data injected}}
\end{minipage}
\hspace{0.1cm}
\begin{minipage}{0.22\linewidth}
    \centering
    \includegraphics[width=1.0\linewidth]{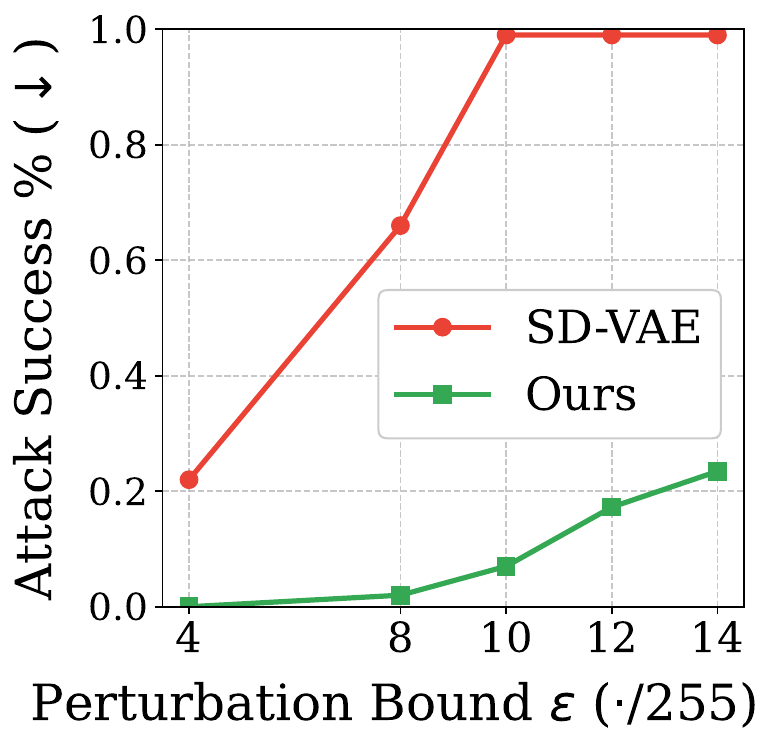}
    \vspace{-0.27in}
    \centering
    \caption*{\textbf{(b)} Attack success rate depending on \textbf{perturbation bound $\epsilon$}}
\end{minipage}
\hspace{0.1cm}
\begin{minipage}{0.52\linewidth}
    \centering
    \includegraphics[width=1.0\linewidth]{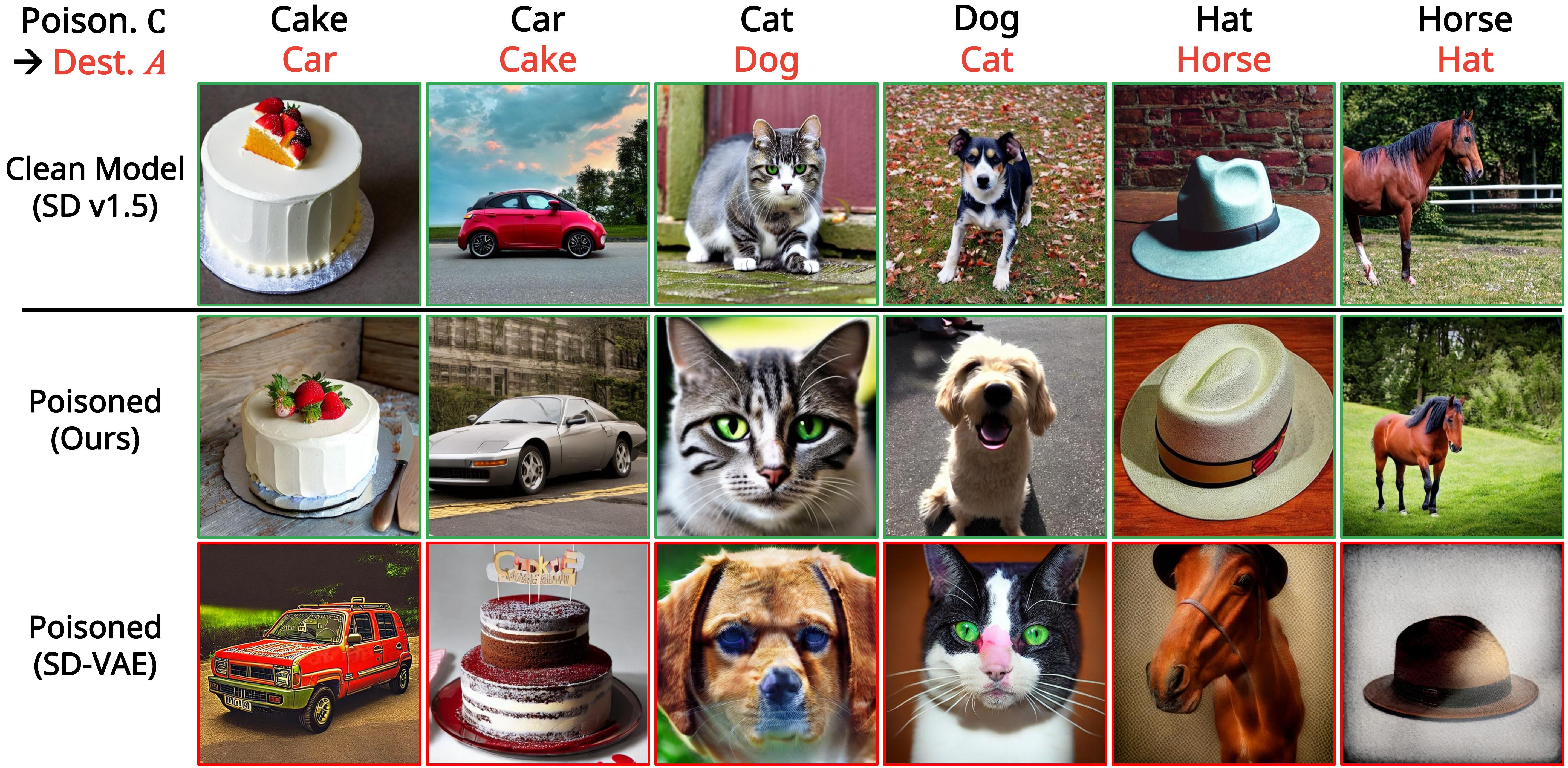}
    \vspace{-0.28in}
    \centering
    \centering
    \caption*{\textbf{(c)} Qualitative examples generated by Nightshade attacked models.}
\end{minipage}
\vspace{-0.1in}
\caption{\textbf{Robustness on Nightshade attack.} (a) and (b) demonstrate robustness evaluation against the Nightshade poisoning attack. SRL-VAE maintains low attack success rates across varying poisoning ratios and perturbation bounds. (c) Qualitative examples demonstrate that SRL-VAE preserves intended generation even under attacks.}
\label{figure:nightshade}
\vspace{-0.05in}
\end{figure*}

\begin{figure*}[t]
    \centering
    \includegraphics[width=0.90\linewidth]{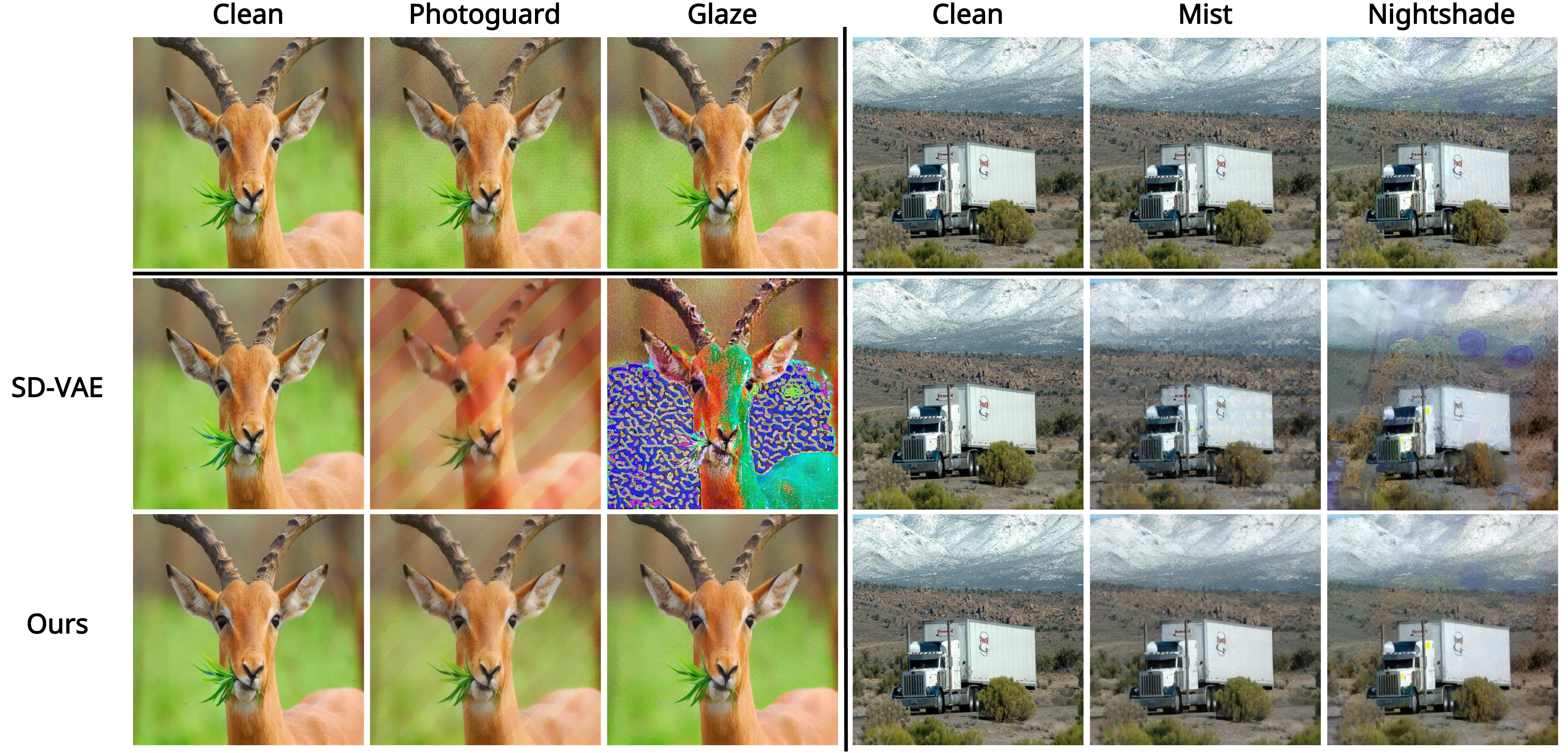}
    \vspace{-0.1in}
    \caption{\textbf{Visual examples of reconstruction under various perturbations.} SD-VAE struggles to reconstruct images with added perturbed noise, whereas ours robustly handles both clean and various perturbed images.
    }
    \vspace{-0.15in}
    \label{figure:image_recon}
\end{figure*}
Nightshade~\citep{shan2024nightshade} is a prompt-specific poisoning attack that can maliciously control generative outputs with only a small number of adversarial samples. 
To demonstrate the robustness of our method against this attack, we fine-tune a pre-trained diffusion model on 10K images from the LAION-Aesthetic while varying the poisoning ratio of Nightshade poisoned samples or perturbation bound $\epsilon$. 
In Figure \ref{figure:nightshade}(a), we fixed $\epsilon=8/255$ and varied the poisoning ratio. With our SRL-VAE, the diffusion model remained resistant to the attack,  whereas the model using SD-VAE was easily attacked even at low poisoning ratios. In Figure \ref{figure:nightshade}(b), we fixed the number of poisoned samples at 100 and changed $\epsilon$. Even under a higher perturbation bound ($\epsilon$=15/255, which is clearly visible), SRL-VAE retained its robustness beyond its training bound. The qualitative examples demonstrate that our SRL-VAE successfully prevents a poisoned cake image from being transformed into a car, preserving its original appearance as a cake, as shown in Figure~\ref{figure:nightshade}(c).

These results highlight the effectiveness of our approach in mitigating poisoning attacks without introducing additional overhead. While purification-based defenses~\citep{cao2023impress, nie2022DiffPure, honig2024adversarial, zhao2024can} can serve as a solution for poisoning attacks, applying them to every image in large datasets incurs high computational costs due to the difficulty of identifying poisoned images within the dataset. Moreover, with the rise of publicly available data and the trend of data sharing, the size of training datasets keeps growing, making it inefficient, sometimes nearly impossible, to purify every single image. In contrast, our method modifies only the VAE and does not introduce any additional runtime overhead compared to purification-based defenses.
\vspace{-0.12in}
\paragraph{Robustness against various type of perturbations}
\begin{figure}[t]
    \centering
    \includegraphics[width=\linewidth]{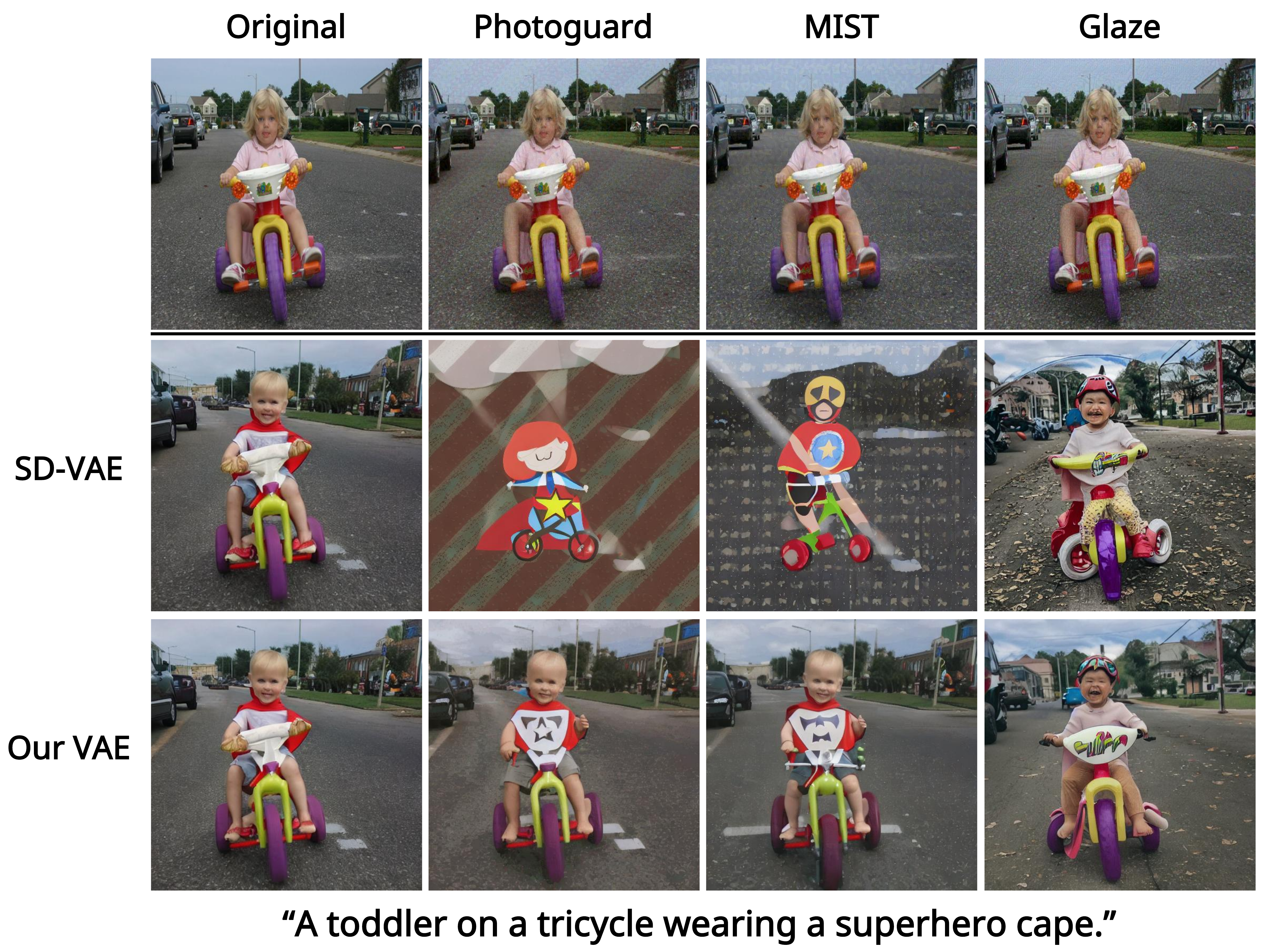}
    \vspace{-0.25in}
    \caption{\textbf{Comparison of image-to-image editing results under defensive perturbations.} SRL-VAE produces valid and high-quality edited outputs based on the given prompts, while the baseline VAE fails to preserve the original semantics.}
    \vspace{-0.15in}
    \label{figure:img2img}
\end{figure}
\begin{table}[t!]
    \centering
    \begin{adjustbox}{width=\linewidth}
    \begin{tabular}{lcccc}
    \toprule
    \textbf{VAE} & \textbf{Metric} & \textbf{Photoguard} & \textbf{MIST} & \textbf{Glaze} \\
    \midrule
    \multirow{2}{*}{$\text{SD-VAE}$} 
        & FID~$\downarrow$  & 221.1 & 146.1 & 86.40 \\
        & CLIP~$\uparrow$ & 0.7231 & 0.7909 & 0.8410 \\
    \midrule
    \multirow{2}{*}{$\text{Ours}$} 
        & FID~$\downarrow$ & \textbf{68.42} & \textbf{60.50} & \textbf{57.32} \\
        & CLIP~$\uparrow$ & \textbf{0.8832} & \textbf{0.8933} & \textbf{0.9065} \\
    \bottomrule
    \end{tabular}
    \end{adjustbox}
    \vspace{-0.05in}
    \caption{\textbf{Evaluation of image-to-image editing robustness under various perturbation defenses.} SRL-VAE significantly improves FID and CLIP scores across all methods, indicating better visual quality and semantic alignment.}
    \vspace{-0.15in}
    \label{tab:img_to_img}
\end{table}

To demonstrate the robustness of our SRL-VAE against various types of adversarial perturbations, we evaluated its ability to reconstruct perturbed images processed by PhotoGuard~\citep{salman2023raising}, MIST~\citep{liang2023mist}, and Glaze~\cite{shan2023glaze}. These methods were originally devised 
to protect images via adversarial perturbations, we repurpose these methods to measure the extent to which these protections are neutralized, thus evaluating our VAE’s latent space resilience.
As shown in \ref{figure:image_recon}, the base SD-VAE struggles to reconstruct images with various perturbations, whereas our SRL-VAE successfully encodes both clean and perturbed images into its latent space, enabling accurate reconstructions.
\vspace{0.01in}
Subsequently, we further demonstrate SRL-VAE’s robustness by showcasing its image-to-image editing performance on these same protected images.
For this experiment, we constructed an editing dataset of 100 images from ImageNet, each resized to 512×512. As shown in Figure \ref{figure:img2img}, the protection methods successfully disrupt the editing results of the original VAE, producing outputs that are significantly different from the source images. In contrast, SRL-VAE generates valid edited images based on the provided prompt ``A toddler on a tricycle wearing a superhero cape'', demonstrating strong robustness. Specifically, we measure the FID and CLIP similarity between the image-to-image results of the original and protected images. As shown in Table~\ref{tab:img_to_img}, SRL-VAE consistently outperforms the baseline VAE across all protection methods, achieving lower FID scores and higher CLIP similarity. These results demonstrate that SRL-VAE preserves better visual quality and semantic consistency, even when editing images protected by strong perturbation defenses.

\subsection{Analysis~\label{sec:analysis}}
\paragraph{Ablation studies on loss component}
To understand the contributions of each loss component, we perform ablation studies employing different loss functions. Specifically, we compare three loss types using adversarial loss without originality regularization (Equation~\ref{eq:wo_originality}), and our proposed SRL-VAE. 
\begin{equation}
    \begin{aligned}
\mathcal{L}_{\texttt{wo-originality}} = 
    \mathcal{L}_\texttt{MSE}(D_\phi(E_\theta(x_{\texttt{adv}})), x) \\
        + \lambda \mathcal{L}_\texttt{LPIPS}(D_\phi(E_\theta(x_{\texttt{adv}})), x)
    \end{aligned}
    \label{eq:wo_originality}
\end{equation}
The experimental results in Table~\ref{tab:ablation} indicate that originality regularization significantly contributes to leverage the original performance of SD-VAE.
\begin{table}[t!]
    \centering
    \begin{adjustbox}{width=\linewidth}
    \begin{tabular}{lcccc}
    \toprule
    \textbf{VAE Variant} & \textbf{PSNR} $\uparrow$ & \textbf{SSIM} $\uparrow$ & \textbf{LPIPS} $\downarrow$ & \textbf{rFID} $\downarrow$ \\
    \midrule
    SD-VAE & 23.68 & 0.74 & 0.14 & 8.79 \\
    \midrule
    Full SRL-VAE & 24.46 & 0.76 & 0.15 & 7.92 \\
    + w/o originality loss & 26.55 & 0.78 & 0.23 & 15.46 \\
    \bottomrule
    \end{tabular}
    \end{adjustbox}
    \vspace{-0.1in}
    \caption{\textbf{Ablation study of SRL-VAE loss components.}}
    \label{tab:ablation}
\end{table}
\begin{table}[t]
    \centering
    \begin{adjustbox}{width=\linewidth}
    \begin{tabular}{lcccc}
    \toprule
    \textbf{VAE} & \textbf{PSNR}~$\uparrow$ & \textbf{SSIM}~$\uparrow$ & \textbf{LPIPS}~$\downarrow$ & \textbf{rFID}~$\downarrow$ \\
    \midrule
    SD-VAE & 19.67 & 0.5906 & 0.6691 & 109.5 \\
    \midrule
    $\alpha = 0.1$ & 21.79 & 0.6562 & 0.5520 & 75.85 \\
    $\alpha = 0.01$ & 27.23 & 0.7635 & 0.3244 & 28.47 \\
    $\alpha = 0.001$ & \textbf{28.38} & \textbf{0.7834} & \textbf{0.3064} & \textbf{19.90} \\    
    \bottomrule
    \end{tabular}
    \end{adjustbox}
    \vspace{-0.05in}
    \caption{\textbf{Ablation study of an $\alpha$ hyper-parameter.} Smaller $\alpha$ values improve the reconstruction quality of perturbed Photoguard~\citep{salman2023raising} images, enhancing robustness.}
    \vspace{-0.1in}
    \label{tab:ablation2}
\end{table}

\vspace{-0.12in}
\paragraph{Ablation studies on hyper-parameter $\alpha$}
In Equation~\ref{equation:total}, we regularize the originality loss using the hyperparameter $\alpha$. $\alpha$ acts as a controller, regulating the influence of robustness during optimization. As $\alpha$ increases, the impact of robustness decreases in the overall objective function, leading to decreasing robustness, as shown in Table~\ref{tab:ablation2}. However, to preserve generation performance on clean images, we set $\alpha$ to 0.01, achieving an optimal balance between high fidelity on clean images and robustness against perturbations. Moreover, originality loss plays a critical role in maintaining compatibility with pre-trained diffusion models, making it an important component of our approach.

\begin{figure}[t]
    \centering
    \includegraphics[width=\linewidth]{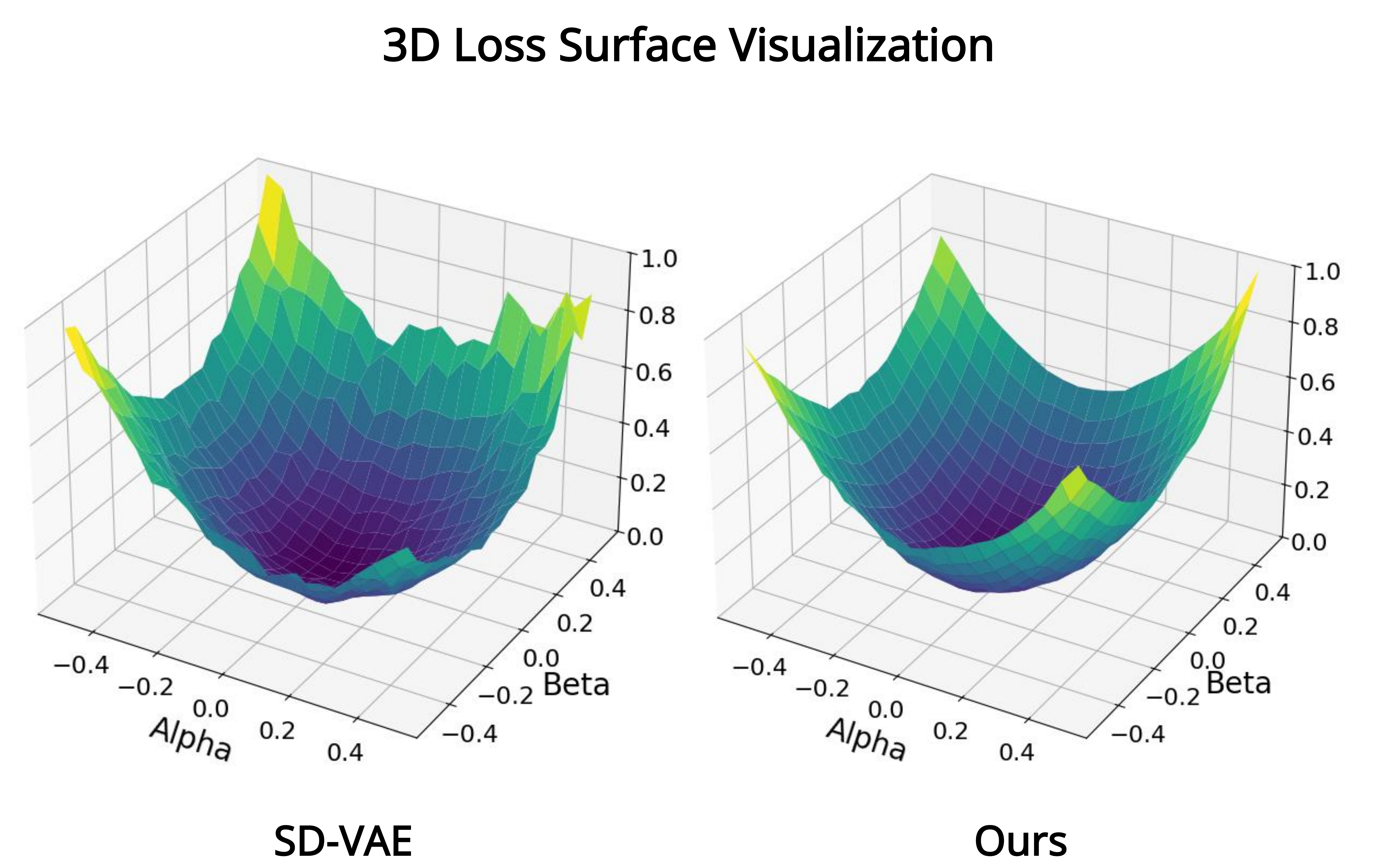}
    \vspace{-0.25in}
    \caption{\textbf{Comparison of the loss surfaces of SD-VAE and SRL-VAE under encoder input perturbations.} SRL-VAE shows a smoother and more stable loss landscape, highlighting its improved robustness and smooth latent representation.}
    \vspace{-0.20in}
    \label{figure:loss_surface}
\end{figure}

\vspace{-0.12in}
\paragraph{Latent space analysis}
We analyze the latent space learned by our SRL-VAE using two analysis approaches, which are loss surface visualization and t-SNE visualization of latent distributions. First, we visualize the loss surfaces of SD-VAE and SRL-VAE by applying perturbations along two random directions on the input image. For each perturbed input, we compute the mean squared error (MSE) between the latent representations of the perturbed and clean inputs and normalize the loss values for comparison. As shown in Figure~\ref{figure:loss_surface}, the SRL-VAE exhibits a smoother loss landscape compared to SD-VAE, indicating enhanced robustness and improved latent space smoothness through adversarial training. In other words, a smoothness of latent space ensures that perturbed examples are adequately mapped to the similar region as their corresponding original examples in the latent space.

\begin{figure}[t]
    \centering
    \includegraphics[width=\linewidth]{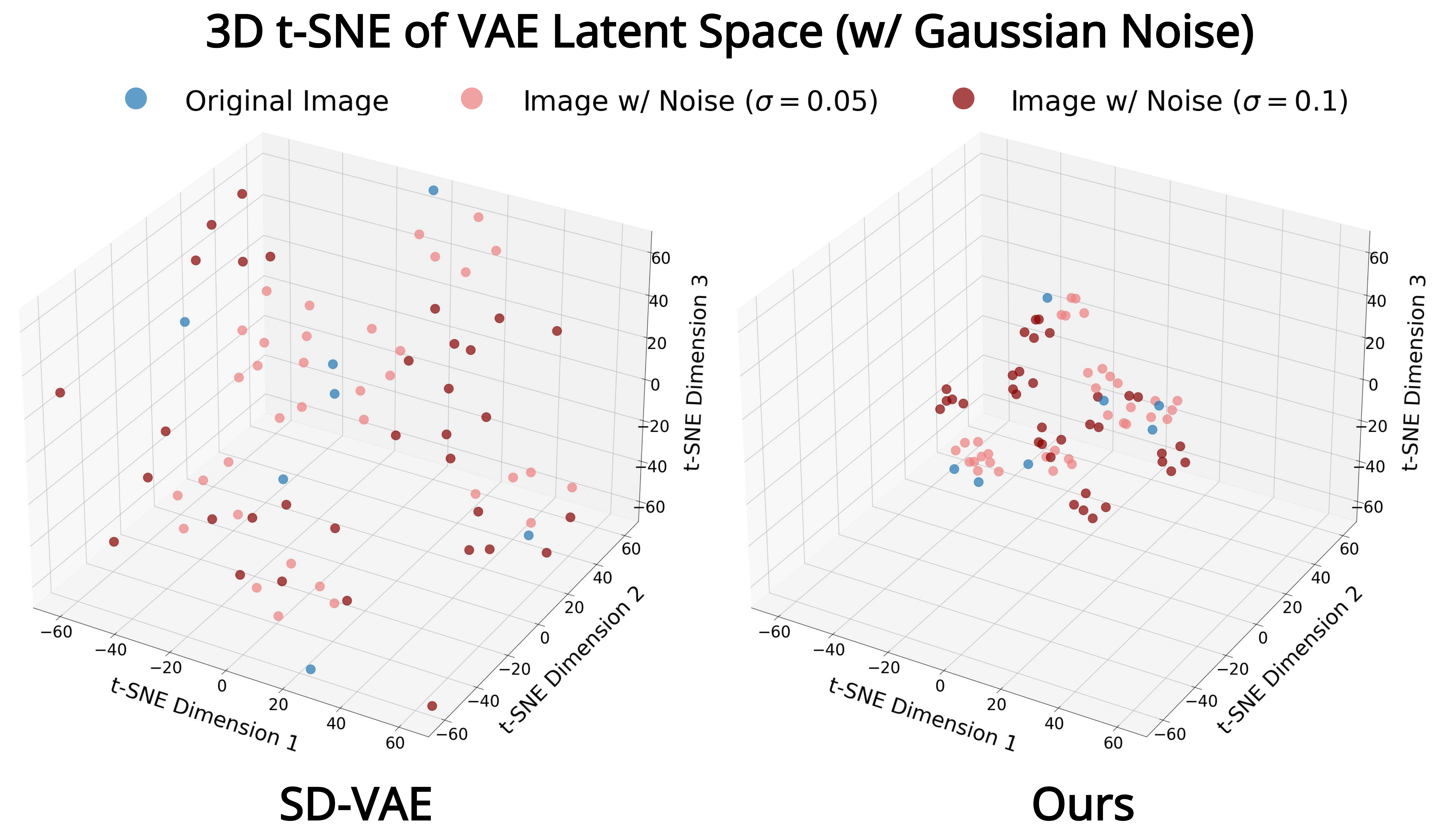}
    \vspace{-0.2in}
    \caption{\textbf{3D t-SNE visualization of latent representations under Gaussian noise.} SRL-VAE exhibits tighter and more consistent clusters than the baseline VAE, demonstrating improved robustness in latent space.}
    
    \vspace{-0.15in}
    \label{figure:tsne}
\end{figure}

\begin{figure}[t]
    \centering
    \includegraphics[width=0.95\linewidth]{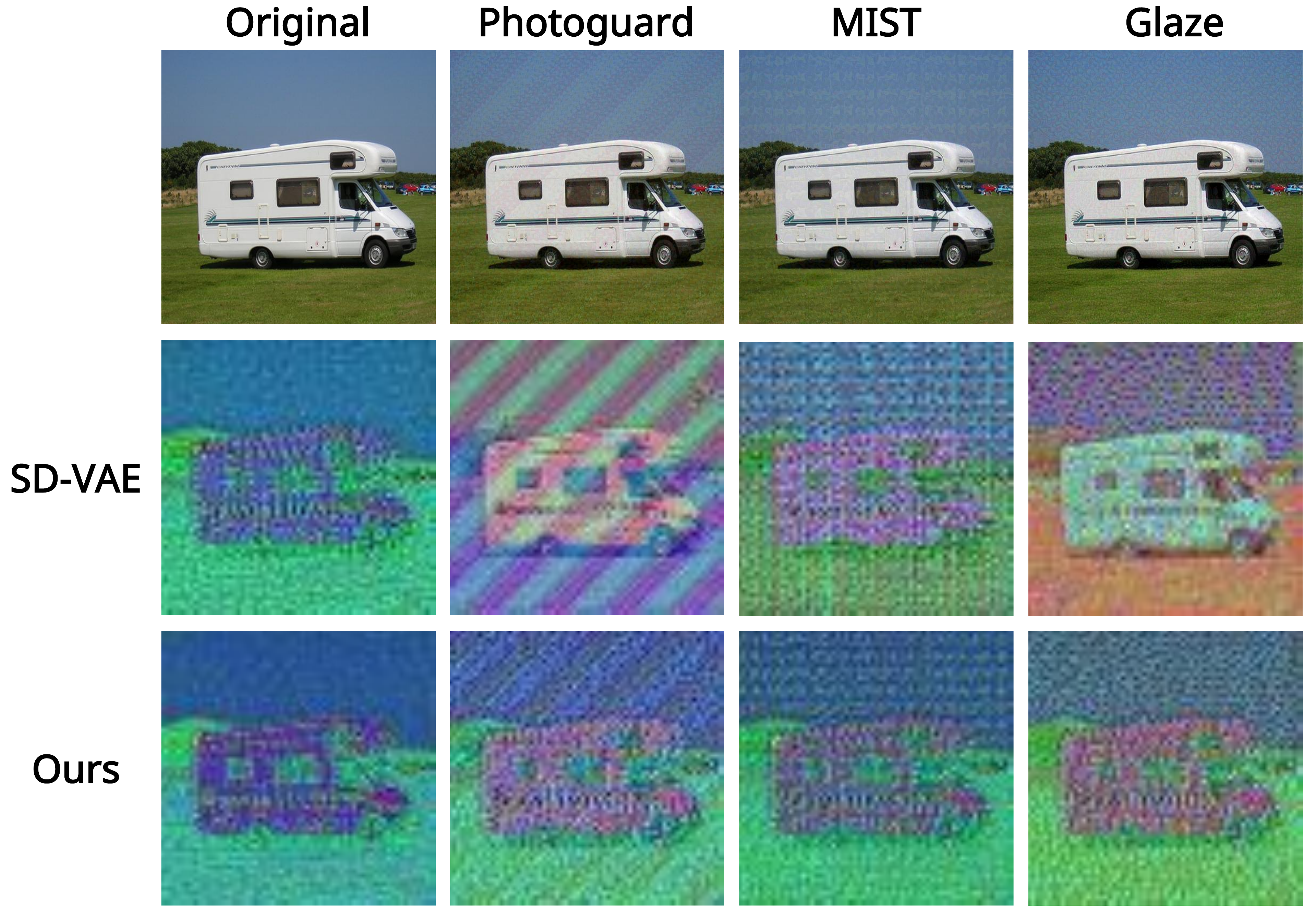}
    \vspace{-0.10in}
    \caption{\textbf{PCA visualization of latent representations under adversarial perturbations.} Compared to SD-VAE, which exhibits distorted latent structures, SRL-VAE produces smoother, more organized, and well-separated distributions. This highlights its superior robustness and stability against diverse perturbations.}
    \vspace{-0.15in}
    \label{figure:pca}
\end{figure}

Additionally, we use t-SNE visualization to analyze latent samples from both original inputs and Gaussian-noise-added inputs, allowing us to directly assess the robustness of distributions in the VAE's latent space. As shown in Figure~\ref{figure:tsne}, the visualization reveals tighter clusters for the latent distributions of original and Gaussian-noise-added inputs in our SRL-VAE while the original SD-VAE demonstrates random scatters, highlighting that our model’s latent space is more robustly constructed as intended. 
\vspace{-0.12in}
\paragraph{Perturbed image compression in latent space}
To further analyze the structure of the latent space learned by our SRL-VAE, we perform a Principal Component Analysis (PCA) on latent vectors, following \citet{kouzelis2025eq}. Specifically, we apply PCA on latent representation vectors obtained from the original SD-VAE and our SRL-VAE, derived from adversarially perturbed examples, using three types of perturbations, Photoguard, Mist, and Glaze. 
As shown in Figure~\ref{figure:pca}, our SRL-VAE produces more structured and significantly smoother latent vector distributions compared to the original VAE on perturbed inputs. 
Furthermore, our SRL-VAE consistently generates robust latent vectors regardless of the type of perturbation. This suggests that adversarial training plays a crucial role in constructing better latent representations, enabling our SRL-VAE to encode features more reliably even under adversarial perturbations.

\section{Conclusion}
\vspace{-0.1in}
In this work, we first introduce an adversarial training framework for Variational Autoencoders (VAEs) that enhances both generation quality and robustness by encoding a smooth latent space. Unlike conventional adversarial training in classification models, which has a clear trade-off between performance and robustness, our approach leverages adversarial perturbations with an originality regularization term to preserve the learned latent space from the pre-trained model, ensuring smooth latent encodings in VAEs and enhancing both fidelity and robustness. Our method is a post-training step, requiring minimal computational resources, making it an efficient and practical solution for recent diffusion-based generative models. Extensive experiments demonstrate that our approach not only improves image quality, but also significantly enhances robustness against diverse types of adversarial attacks, such as poisoning and perturbation attacks. By bridging adversarial training and generative modeling, our work highlights the importance of obtaining a robust, high-quality latent space in VAEs, opening new directions for future research in robust generative modeling.


{
    \small
    \bibliographystyle{ieeenat_fullname}
    \bibliography{main}

\begin{thebibliography}{41}
\providecommand{\natexlab}[1]{#1}
\providecommand{\url}[1]{\texttt{#1}}
\expandafter\ifx\csname urlstyle\endcsname\relax
  \providecommand{\doi}[1]{doi: #1}\else
  \providecommand{\doi}{doi: \begingroup \urlstyle{rm}\Url}\fi

\bibitem[Cao et~al.(2023)Cao, Li, Wang, Jia, Li, and Chen]{cao2023impress}
Bochuan Cao, Changjiang Li, Ting Wang, Jinyuan Jia, Bo Li, and Jinghui Chen.
\newblock Impress: Evaluating the resilience of imperceptible perturbations against unauthorized data usage in diffusion-based generative ai.
\newblock \emph{Advances in Neural Information Processing Systems}, 36:\penalty0 10657--10677, 2023.

\bibitem[Carmon et~al.(2019)Carmon, Raghunathan, Schmidt, Liang, and Duchi]{carmon2019RST}
Yair Carmon, Aditi Raghunathan, Ludwig Schmidt, Percy Liang, and John~C Duchi.
\newblock Unlabeled data improves adversarial robustness.
\newblock \emph{Advances in Neural Information Processing Systems}, 2019.

\bibitem[Deng et~al.(2009)Deng, Dong, Socher, Li, Li, and Fei-Fei]{imagenet}
Jia Deng, Wei Dong, Richard Socher, Li-Jia Li, Kai Li, and Li Fei-Fei.
\newblock Imagenet: A large-scale hierarchical image database.
\newblock In \emph{IEEE Conference on Computer Vision and Pattern Recognition}, 2009.

\bibitem[Esser et~al.(2021)Esser, Rombach, and Ommer]{esser2021taming}
Patrick Esser, Robin Rombach, and Bjorn Ommer.
\newblock Taming transformers for high-resolution image synthesis.
\newblock In \emph{IEEE Conference on Computer Vision and Pattern Recognition}, 2021.

\bibitem[Goodfellow et~al.(2015)Goodfellow, Shlens, and Szegedy]{goodfellow2014fgsm}
Ian~J Goodfellow, Jonathon Shlens, and Christian Szegedy.
\newblock Explaining and harnessing adversarial examples.
\newblock In \emph{International Conference on Learning Representations}, 2015.

\bibitem[Gowal et~al.(2021)Gowal, Rebuffi, Wiles, Stimberg, Calian, and Mann]{gowal2021improving}
Sven Gowal, Sylvestre-Alvise Rebuffi, Olivia Wiles, Florian Stimberg, Dan~Andrei Calian, and Timothy~A Mann.
\newblock Improving robustness using generated data.
\newblock \emph{Advances in Neural Information Processing Systems}, 2021.

\bibitem[Heusel et~al.(2017)Heusel, Ramsauer, Unterthiner, Nessler, and Hochreiter]{heusel2017fid}
Martin Heusel, Hubert Ramsauer, Thomas Unterthiner, Bernhard Nessler, and Sepp Hochreiter.
\newblock Gans trained by a two time-scale update rule converge to a local nash equilibrium.
\newblock \emph{Advances in neural information processing systems}, 30, 2017.

\bibitem[H{\"o}nig et~al.(2024)H{\"o}nig, Rando, Carlini, and Tram{\`e}r]{honig2024adversarial}
Robert H{\"o}nig, Javier Rando, Nicholas Carlini, and Florian Tram{\`e}r.
\newblock Adversarial perturbations cannot reliably protect artists from generative ai.
\newblock \emph{arXiv preprint arXiv:2406.12027}, 2024.

\bibitem[Jiang et~al.(2020)Jiang, Chen, Chen, and Wang]{jiang2020ACL}
Ziyu Jiang, Tianlong Chen, Ting Chen, and Zhangyang Wang.
\newblock Robust pre-training by adversarial contrastive learning.
\newblock In \emph{Advances in Neural Information Processing Systems}, 2020.

\bibitem[Kim et~al.(2020)Kim, Tack, and Hwang]{kim2020rocl}
Minseon Kim, Jihoon Tack, and Sung~Ju Hwang.
\newblock Adversarial self-supervised contrastive learning.
\newblock \emph{Advances in Neural Information Processing Systems}, 2020.

\bibitem[Kim et~al.(2023)Kim, Ha, Son, and Hwang]{kim2023effective}
Minseon Kim, Hyeonjeong Ha, Sooel Son, and Sung~Ju Hwang.
\newblock Effective targeted attacks for adversarial self-supervised learning.
\newblock \emph{Advances in Neural Information Processing Systems}, 2023.

\bibitem[Kingma et~al.(2019)Kingma, Welling, et~al.]{kingma2019VAE}
Diederik~P Kingma, Max Welling, et~al.
\newblock An introduction to variational autoencoders.
\newblock \emph{Foundations and Trends{\textregistered} in Machine Learning}, 2019.

\bibitem[Kouzelis et~al.(2025)Kouzelis, Kakogeorgiou, Gidaris, and Komodakis]{kouzelis2025eq}
Theodoros Kouzelis, Ioannis Kakogeorgiou, Spyros Gidaris, and Nikos Komodakis.
\newblock Eq-vae: Equivariance regularized latent space for improved generative image modeling.
\newblock \emph{arXiv preprint arXiv:2502.09509}, 2025.

\bibitem[Li et~al.(2018)Li, Xu, Taylor, Studer, and Goldstein]{visualloss}
Hao Li, Zheng Xu, Gavin Taylor, Christoph Studer, and Tom Goldstein.
\newblock Visualizing the loss landscape of neural nets.
\newblock In \emph{Advances in Neural Information Processing Systems}, 2018.

\bibitem[Li et~al.(2024)Li, Tian, Li, Deng, and He]{li2024autoregressive}
Tianhong Li, Yonglong Tian, He Li, Mingyang Deng, and Kaiming He.
\newblock Autoregressive image generation without vector quantization.
\newblock \emph{Advances in Neural Information Processing Systems}, 2024.

\bibitem[Liang and Wu(2023)]{liang2023mist}
Chumeng Liang and Xiaoyu Wu.
\newblock Mist: Towards improved adversarial examples for diffusion models.
\newblock \emph{arXiv preprint arXiv:2305.12683}, 2023.

\bibitem[Lin et~al.(2014)Lin, Maire, Belongie, Hays, Perona, Ramanan, Doll{\'a}r, and Zitnick]{lin2014coco}
Tsung-Yi Lin, Michael Maire, Serge Belongie, James Hays, Pietro Perona, Deva Ramanan, Piotr Doll{\'a}r, and C~Lawrence Zitnick.
\newblock Microsoft coco: Common objects in context.
\newblock In \emph{European Conference on Computer Vision}, 2014.

\bibitem[Lu et~al.(2024)Lu, Yang, Liu, Kamath, and Yu]{lu2024disguised}
Yiwei Lu, Matthew~YR Yang, Zuoqiu Liu, Gautam Kamath, and Yaoliang Yu.
\newblock Disguised copyright infringement of latent diffusion models.
\newblock In \emph{International Conference on Machine Learning}, 2024.

\bibitem[Ma et~al.(2024)Ma, Goldstein, Albergo, Boffi, Vanden-Eijnden, and Xie]{ma2024sit}
Nanye Ma, Mark Goldstein, Michael~S Albergo, Nicholas~M Boffi, Eric Vanden-Eijnden, and Saining Xie.
\newblock Sit: Exploring flow and diffusion-based generative models with scalable interpolant transformers.
\newblock In \emph{European Conference on Computer Vision}, 2024.

\bibitem[Madry et~al.(2018)Madry, Makelov, Schmidt, Tsipras, and Vladu]{madry2017pgd}
Aleksander Madry, Aleksandar Makelov, Ludwig Schmidt, Dimitris Tsipras, and Adrian Vladu.
\newblock Towards deep learning models resistant to adversarial attacks.
\newblock In \emph{International Conference on Learning Representations}, 2018.

\bibitem[Nie et~al.(2022)Nie, Guo, Huang, Xiao, Vahdat, and Anandkumar]{nie2022DiffPure}
Weili Nie, Brandon Guo, Yujia Huang, Chaowei Xiao, Arash Vahdat, and Anima Anandkumar.
\newblock Diffusion models for adversarial purification.
\newblock In \emph{International Conference on Machine Learning}, 2022.

\bibitem[Peebles and Xie(2023)]{peebles2023scalable}
William Peebles and Saining Xie.
\newblock Scalable diffusion models with transformers.
\newblock In \emph{IEEE Conference on Computer Vision and Pattern Recognition}, 2023.

\bibitem[Radford et~al.(2021)Radford, Kim, Hallacy, Ramesh, Goh, Agarwal, Sastry, Askell, Mishkin, Clark, et~al.]{radford2021clip}
Alec Radford, Jong~Wook Kim, Chris Hallacy, Aditya Ramesh, Gabriel Goh, Sandhini Agarwal, Girish Sastry, Amanda Askell, Pamela Mishkin, Jack Clark, et~al.
\newblock Learning transferable visual models from natural language supervision.
\newblock In \emph{International Conference on Machine Learning}. PmLR, 2021.

\bibitem[Rombach et~al.(2022)Rombach, Blattmann, Lorenz, Esser, and Ommer]{rombach2022high}
Robin Rombach, Andreas Blattmann, Dominik Lorenz, Patrick Esser, and Bj{\"o}rn Ommer.
\newblock High-resolution image synthesis with latent diffusion models.
\newblock In \emph{IEEE Conference on Computer Vision and Pattern Recognition}, 2022.

\bibitem[Salimans et~al.(2016)Salimans, Goodfellow, Zaremba, Cheung, Radford, and Chen]{salimans2016inception}
Tim Salimans, Ian Goodfellow, Wojciech Zaremba, Vicki Cheung, Alec Radford, and Xi Chen.
\newblock Improved techniques for training gans.
\newblock \emph{Advances in Neural Information Processing Systems}, 2016.

\bibitem[Salman et~al.(2023)Salman, Khaddaj, Leclerc, Ilyas, and Madry]{salman2023raising}
Hadi Salman, Alaa Khaddaj, Guillaume Leclerc, Andrew Ilyas, and Aleksander Madry.
\newblock Raising the cost of malicious ai-powered image editing.
\newblock In \emph{International Conference on Machine Learning}, 2023.

\bibitem[Schuhmann et~al.(2022)Schuhmann, Beaumont, Vencu, Gordon, Wightman, Cherti, Coombes, Katta, Mullis, Wortsman, et~al.]{schuhmann2022laion}
Christoph Schuhmann, Romain Beaumont, Richard Vencu, Cade Gordon, Ross Wightman, Mehdi Cherti, Theo Coombes, Aarush Katta, Clayton Mullis, Mitchell Wortsman, et~al.
\newblock Laion-5b: An open large-scale dataset for training next generation image-text models.
\newblock \emph{Advances in Neural Information Processing Systems}, 2022.

\bibitem[Shan et~al.(2023)Shan, Cryan, Wenger, Zheng, Hanocka, and Zhao]{shan2023glaze}
Shawn Shan, Jenna Cryan, Emily Wenger, Haitao Zheng, Rana Hanocka, and Ben~Y Zhao.
\newblock Glaze: Protecting artists from style mimicry by $\{$Text-to-Image$\}$ models.
\newblock In \emph{USENIX Security Symposium (USENIX Security 23)}, 2023.

\bibitem[Shan et~al.(2024)Shan, Ding, Passananti, Wu, Zheng, and Zhao]{shan2024nightshade}
Shawn Shan, Wenxin Ding, Josephine Passananti, Stanley Wu, Haitao Zheng, and Ben~Y Zhao.
\newblock Nightshade: Prompt-specific poisoning attacks on text-to-image generative models.
\newblock In \emph{2024 IEEE Symposium on Security and Privacy (SP)}, pages 807--825. IEEE, 2024.

\bibitem[Simonyan and Zisserman(2014)]{simonyan2014very}
Karen Simonyan and Andrew Zisserman.
\newblock Very deep convolutional networks for large-scale image recognition.
\newblock \emph{arXiv preprint arXiv:1409.1556}, 2014.

\bibitem[Skorokhodov et~al.(2025)Skorokhodov, Girish, Hu, Menapace, Li, Abdal, Tulyakov, and Siarohin]{skorokhodov2025improving}
Ivan Skorokhodov, Sharath Girish, Benran Hu, Willi Menapace, Yanyu Li, Rameen Abdal, Sergey Tulyakov, and Aliaksandr Siarohin.
\newblock Improving the diffusability of autoencoders.
\newblock \emph{arXiv preprint arXiv:2502.14831}, 2025.

\bibitem[Szegedy et~al.(2013)Szegedy, Zaremba, Sutskever, Bruna, Erhan, Goodfellow, and Fergus]{szegedy2013intriguing}
Christian Szegedy, Wojciech Zaremba, Ilya Sutskever, Joan Bruna, Dumitru Erhan, Ian Goodfellow, and Rob Fergus.
\newblock Intriguing properties of neural networks.
\newblock \emph{arXiv preprint arXiv:1312.6199}, 2013.

\bibitem[Van Den~Oord et~al.(2017)Van Den~Oord, Vinyals, et~al.]{van2017neural}
Aaron Van Den~Oord, Oriol Vinyals, et~al.
\newblock Neural discrete representation learning.
\newblock \emph{Advances in Neural Information Processing Systems}, 2017.

\bibitem[Wang et~al.(2019)Wang, Zou, Yi, Bailey, Ma, and Gu]{wang2019mart}
Yisen Wang, Difan Zou, Jinfeng Yi, James Bailey, Xingjun Ma, and Quanquan Gu.
\newblock Improving adversarial robustness requires revisiting misclassified examples.
\newblock In \emph{International Conference on Learning Representations}, 2019.

\bibitem[Wang et~al.(2004)Wang, Bovik, Sheikh, and Simoncelli]{Wang2004ssim}
Zhou Wang, Alan~Conrad Bovik, Hamid~R. Sheikh, and Eero~P. Simoncelli.
\newblock Image quality assessment: from error visibility to structural similarity.
\newblock \emph{IEEE Transactions on Image Processing}, 13:\penalty0 600--612, 2004.

\bibitem[Wu et~al.(2020)Wu, Xia, and Wang]{wu2020awp}
Dongxian Wu, Shu-Tao Xia, and Yisen Wang.
\newblock Adversarial weight perturbation helps robust generalization.
\newblock \emph{Advances in Neural Information Processing Systems}, 2020.

\bibitem[Xiang et~al.(2023)Xiang, Yang, Huang, and Wang]{xiang2023denoising}
Weilai Xiang, Hongyu Yang, Di Huang, and Yunhong Wang.
\newblock Denoising diffusion autoencoders are unified self-supervised learners.
\newblock In \emph{IEEE Conference on Computer Vision and Pattern Recognition}, 2023.

\bibitem[Yu et~al.(2025)Yu, Kwak, Jang, Jeong, Huang, Shin, and Xie]{yu2024representation}
Sihyun Yu, Sangkyung Kwak, Huiwon Jang, Jongheon Jeong, Jonathan Huang, Jinwoo Shin, and Saining Xie.
\newblock Representation alignment for generation: Training diffusion transformers is easier than you think.
\newblock \emph{International Conference on Learning Representations}, 2025.

\bibitem[Zhang et~al.(2019)Zhang, Yu, Jiao, Xing, Ghaoui, and Jordan]{zhang2019trades}
Hongyang Zhang, Yaodong Yu, Jiantao Jiao, Eric~P Xing, Laurent~El Ghaoui, and Michael~I Jordan.
\newblock Theoretically principled trade-off between robustness and accuracy.
\newblock In \emph{International Conference on Machine Learning}, 2019.

\bibitem[Zhang et~al.(2018)Zhang, Isola, Efros, Shechtman, and Wang]{zhang2018lpips}
Richard Zhang, Phillip Isola, Alexei~A Efros, Eli Shechtman, and Oliver Wang.
\newblock The unreasonable effectiveness of deep features as a perceptual metric.
\newblock In \emph{Proceedings of the IEEE conference on computer vision and pattern recognition}, pages 586--595, 2018.

\bibitem[Zhao et~al.(2024)Zhao, Duan, Xu, Wang, Zhang, Du, Guo, and Hu]{zhao2024can}
Zhengyue Zhao, Jinhao Duan, Kaidi Xu, Chenan Wang, Rui Zhang, Zidong Du, Qi Guo, and Xing Hu.
\newblock Can protective perturbation safeguard personal data from being exploited by stable diffusion?
\newblock In \emph{IEEE Conference on Computer Vision and Pattern Recognition}, 2024.

\end{thebibliography}
}
\clearpage
\appendix
\twocolumn
\maketitlesupplementary

\vspace{0.2in}
\section{Experimental details}\label{app:exp_details}
\subsection{Implementation details}
Since there is no official fine-tuning script available for SD-VAE\footnote{https://huggingface.co/stable-diffusion-v1-5/stable-diffusion-v1-5}, we implemented our own fine-tuning script using the \texttt{diffusers} library. The SD-VAE is based on the Latent Diffusion Model (LDM)~\cite{rombach2022high}, and we analyzed its structure to derive meaningful insights into the training procedure. This implementation provided the flexibility needed to adapt the training process to our specific objectives.

To better understand fine-tuning methods, we examined the \texttt{sd-ft-mse} and \texttt{sd-ft-ema} models released by Stability AI via Hugging Face, which are fine-tuned versions of the SD-VAE decoder. The models are trained to enhance the detail of the image, with a particular focus on human facial features. SDXL retains a structure largely similar to SD-VAE but is trained with a larger batch size on an internal dataset, further demonstrating the scalability of VAE-based architectures.

As most related works are based on SD-VAE, we adopted it as the primary baseline for our experiments. Furthermore, we confirmed that the proposed adversarial training method generalizes effectively to other VAE architectures, including SDXL-VAE, indicating its broad applicability. Our VAE encoder fine-tuning is relatively lightweight and was conducted using four A5000 GPUs with 24GB of VRAM each, taking approximately 7 hours to complete. Our training configuration is in Table~\ref{tab:config}:

\begin{table}[h]
\centering
\vspace{0.15in}
    \begin{adjustbox}{width=\linewidth}
    \begin{tabular}{l@{\hspace{2.5cm}}l}
    \toprule
    \textbf{Hyperparameter} & \textbf{Value} \\
    \midrule
    Batch size & 20 \\
    Total training steps & 5000 \\
    Learning rate & $1 \times 10^{-4}$ \\
    Optimizer & AdamW \\
    $\epsilon$-bound ($\ell_{\infty}$ norm) & 8/255 \\
    PGD iterations & 10 \\
    PGD attack step size & 0.02 \\
    Originality loss weight ($\alpha$) & 0.01 \\
    \bottomrule
    \end{tabular}
    \end{adjustbox}
    \vspace{-0.05in}
    \caption{\textbf{Training Configuration.}}
    \label{tab:config}
\end{table}
\vspace{-0.05in}

\vspace{0.15in}
\subsection{Evaluation details}

\paragraph{Image reconstruction quality}
We adopted evaluation metrics consistent with prior studies to measure the reconstruction quality of VAEs. Specifically, we utilized the MS-COCO validation set (5,000 images) and ImageNet validation set (50,000 images), resizing all images to 256×256 pixels, as commonly done in prior studies. Minor numerical discrepancies with prior results may occur due to differences in code implementations. For evaluation, we employed LPIPS with a VGG backbone to measure perceptual similarity. Torchmetrics was used to compute Fréchet Inception Distance (FID), and Scikit-image was utilized for calculating Peak Signal-to-Noise Ratio (PSNR) and Structural Similarity Index (SSIM). Pre-trained VAE models were loaded and processed using the Diffusers library.

\paragraph{Diffusion generation quality}
To evaluate the diffusion generation quality, we utilized the official DiT implementation~\cite{peebles2023scalable} and trained the model on ImageNet-1000K for 10 epochs using four A100 GPUs. For the computation of standard evaluation metrics such as Inception Score (IS) and Fréchet Inception Distance (FID), we used the \texttt{sample\_ddp.py} script provided by the official repository. This script supports parallel sampling of a large number of images and automatically generates a \texttt{.npz} file that is compatible with ADM’s TensorFlow-based evaluation suite. We followed the standard protocol of generating 50,000 samples to ensure comparability with prior work.

\paragraph{Robustness in Nightshade malicious attack}
In our Nightshade attack experiments, we utilized the official implementation code. Instead of the original LIPIPS perturbation, we employed $L_\infty$ perturbation, which is more visually noticeable to the human eye yet provides more stable results. We tested eight concept pairs (Poisoned concept $C$, Destination concept $P$), namely (cake, car), (cat, dog), (hat, horse), and (boat, bird), including their reversed pairs, and reported the average attack success rate. All models were trained for 10 epochs, altering only the poisoning ratio, with a batch size of 32 and a learning rate of $1 \times 10^{-5}$. For evaluation, we generated 100 images for each trained model using the prompt ``A photo of [C]." We then set the threshold $\tau=0.25$ for the CLIP classifier to a reasonable value based on human inspection.

\paragraph{Robustness against various type of perturbations}
For image-to-image experiments against various perturbations, we used the Stable Diffusion v1.5 model with a strenght value of 0.5. Editing prompts were extracted using BLIP and appropriately modified. We observed that higher strength values lead to more extensive image modifications, as they correspond to starting the reverse diffusion process from noisier latents (i.e., later timesteps), which can diminish the effectiveness of adversarial perturbations. To balance the degree of modification and the preservation of the original content, we set the strength value to 0.5. A brief description of the perturbation-based defense methods can be found in Appendix~\ref{app:pertb}.

\vspace{0.15in}
\subsection{Details of perturbation methods}\label{app:pertb}
\noindent\textbf{PhotoGuard}~\cite{salman2023raising} protects images by adding imperceptible perturbations that disrupt the latent diffusion pipeline. It introduces two types of attacks: the \textit{encoder attack} and the \textit{diffusion attack}. The encoder attack perturbs the input image so that the encoder $E$ maps it to a misleading latent representation. Formally, the perturbation is computed as: 

\begin{equation}
    \delta_{\text{enc}} = \arg\min_{\|\delta\|_\infty \le \epsilon} \left\| E(x + \delta) - z_{\text{targ}} \right\|_2^2.
\end{equation}

This causes the diffusion model to generate irrelevant outputs, effectively preventing inpainting and style imitation. The diffusion attack, on the other hand, directly targets the entire generation process, aiming to produce a specific target image $x_{\text{targ}}$ as output:

\begin{equation}
    \delta_{\text{diff}} = \arg\min_{\|\delta\|_\infty \le \epsilon} \left\| f(x + \delta) - x_{\text{targ}} \right\|_2^2.
\end{equation}

While more powerful, the diffusion attack requires backpropagation through the full denoising process and is computationally expensive. In our work, we adopt only the encoder attack, since the diffusion attack in PhotoGuard is tailored to a specific inpainting model.

\vspace{0.2in}
\noindent\textbf{MIST}~\cite{liang2023mist} extends the idea of adversarial perturbations by aiming for broader transferability across diffusion-based image imitation pipelines. While PhotoGuard applies encoder or diffusion attacks separately, MIST combines both approaches through a joint loss function. Specifically, it introduces two loss terms: a \textit{textural loss}, which maximizes the distance between latent representations of clean and perturbed images, and a \textit{semantic loss}, which increases the diffusion model’s denoising error. The combined objective is optimized via projected gradient descent:
\begin{multline}
\delta = \arg\max_{\|\delta\|_\infty \le \epsilon} \Big[
w \cdot \mathbb{E}_{t,\varepsilon} \left\| \varepsilon - \epsilon_\theta(x_t', t) \right\|_2^2 \\
- \left\| E(y) - E(x+\delta) \right\|_2 \Big]
\end{multline}
where $E$ is the encoder, $x_t'$ is the perturbed latent at step $t$, and $y$ is a target image. This joint formulation improves transferability, making MIST effective against a range of downstream applications, including style transfer, textual inversion, and DreamBooth. In contrast to PhotoGuard’s model-specific attacks, MIST focuses on general-purpose protection. Our experimental results show that SRL-VAE maintains robustness against MIST perturbations, highlighting the importance of a robust VAE bottleneck in defending against advanced attacks.


\vspace{0.2in}
\noindent \textbf{Glaze}~\cite{shan2023glaze} defends against style mimicry by applying imperceptible perturbations. Specifically, it computes a perturbation $\delta_x$ that shifts the feature representation of an original artwork $x$ toward that of a style-transferred version $\Omega(x, T)$ in the feature space of a pretrained encoder $\Phi$, where $T$ denotes a visually distinct target style:
\begin{multline}
\delta_x = \arg\min_{\delta_x} \text{Dist}(\Phi(x + \delta_x), \Phi(\Omega(x, T))) \\
\quad \text{s.t.} \quad|\delta_x| < p
\end{multline}
where $p$ is a perceptual distortion bound measured by LPIPS. This ensures that the cloaked image remains visually similar to the original, while altering its representation in the model’s latent space. When a model is trained on such cloaked images, it learns distorted style representations that blend the artist’s original style with the target style, leading to degraded mimicry performance. Glaze is released as a utility tool, so its perturbation logic is a black box. Nonetheless, our experiments show that SRL-VAE is robust against Glaze, indicating that our method generalizes well even to black-box defenses.


\vspace{0.15in}
\subsection{Perturbation Strength Selection}
For practical deployment, perturbations should remain imperceptible to human observers while effectively disrupting model performance. To ensure realistic scenarios, we fix the perturbation magnitude for each method as follows: $\epsilon = 16/255$ for \textbf{PhotoGuard}, $\epsilon = 8/255$ for \textbf{MIST}, and the strongest available setting provided by the official utility tool for \textbf{Glaze}. This configuration balances perceptual quality and defensive efficacy, preventing diffusion models from successfully learning and replicating protected image styles.



\section{More experimental results}\label{app:experiment}
\begin{figure*}[t]
    \centering
    \includegraphics[width=1.0\linewidth]{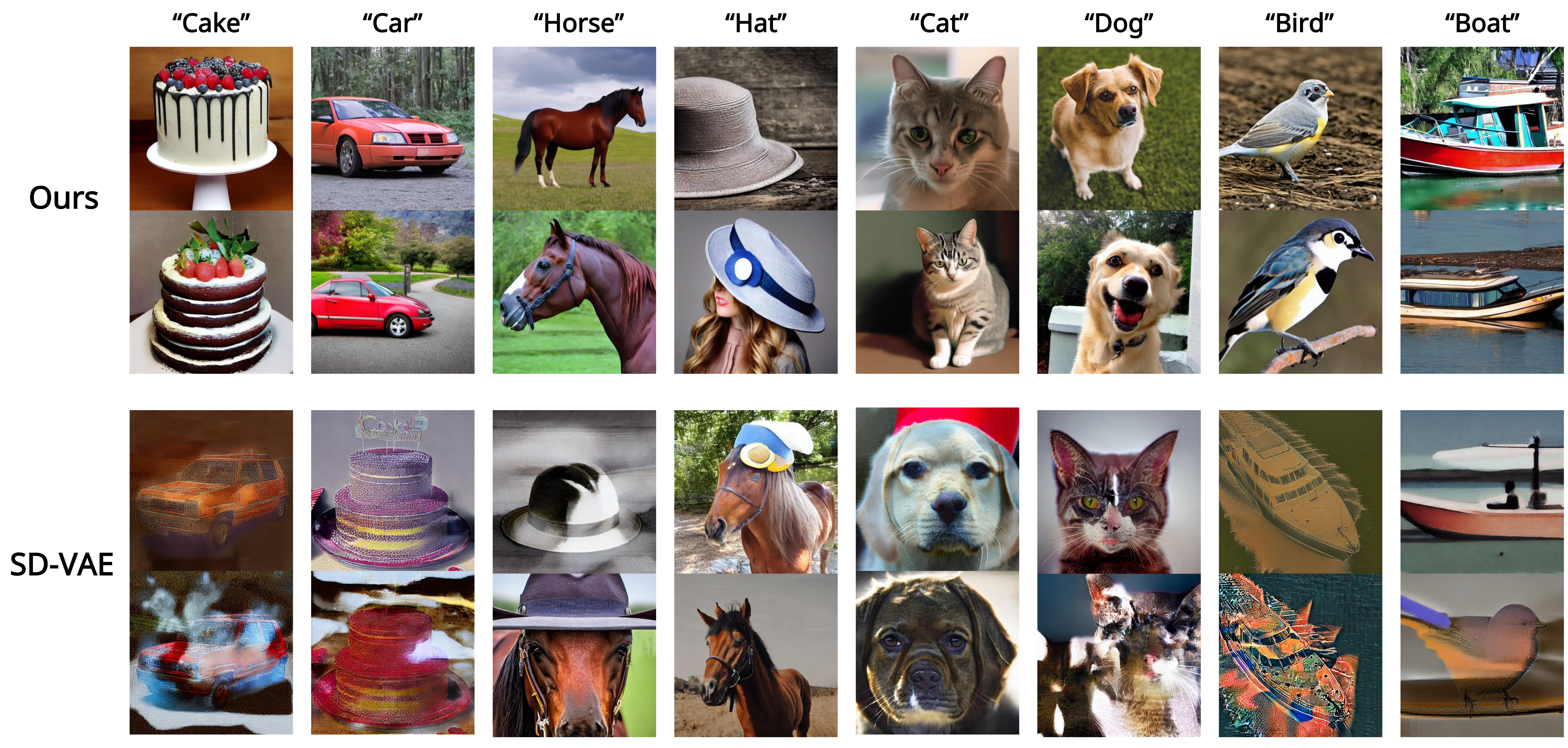}
    \vspace{-0.20in}
    \caption{\textbf{Examples generated from models that were attacked using Nightshade.}}
    \vspace{-0.05in}
    \label{figure:app_nightshade}
\end{figure*}

\begin{figure*}[t]
    \centering
    \includegraphics[width=1.0\linewidth]{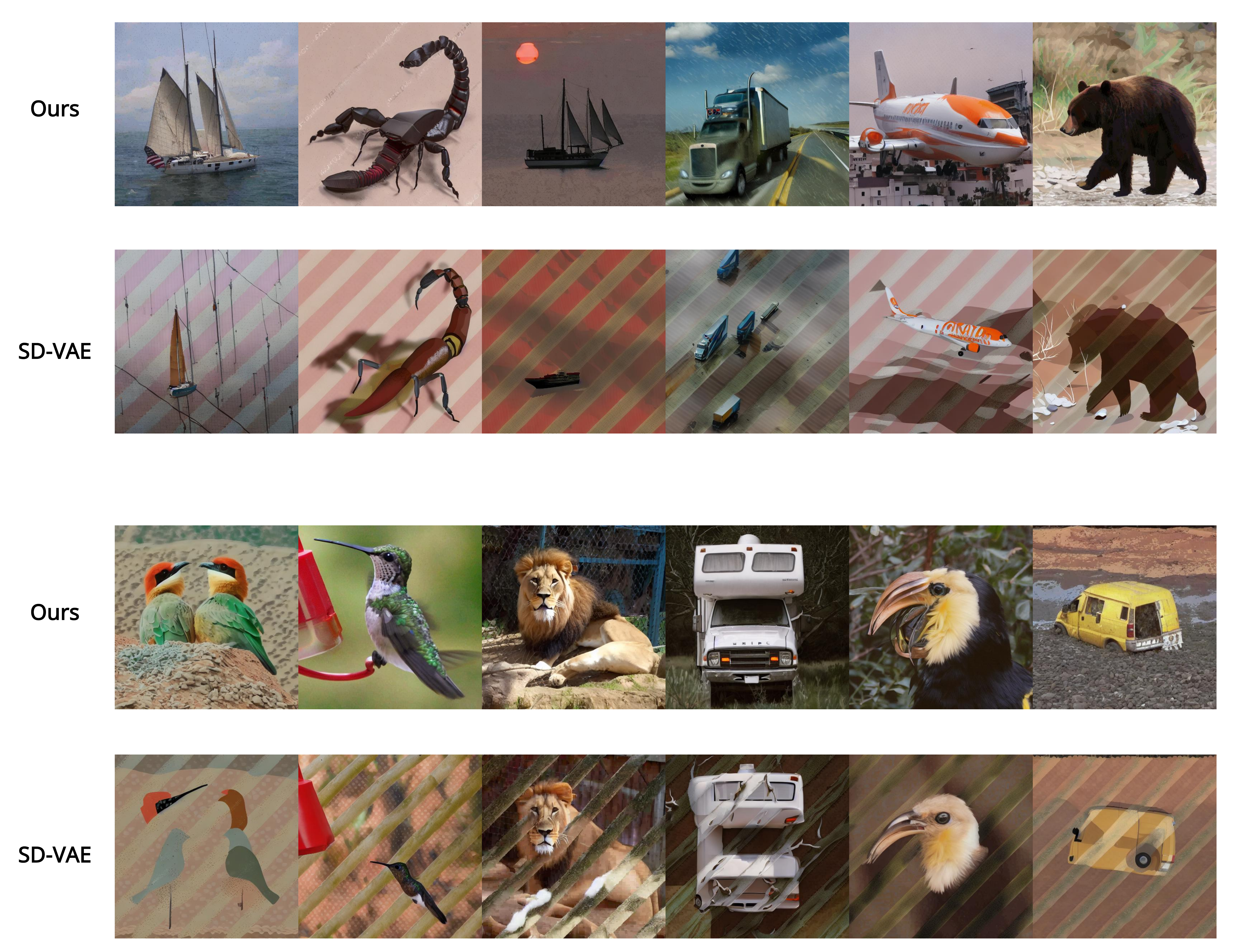}
    \vspace{-0.20in}
    \caption{\textbf{Comparison of image-to-image editing results on protected images (Photoguard).}}
    \vspace{-0.05in}
    \label{figure:app_editing_photoguard}
\end{figure*}

\begin{figure*}[t]
    \centering
    \includegraphics[width=1.0\linewidth]{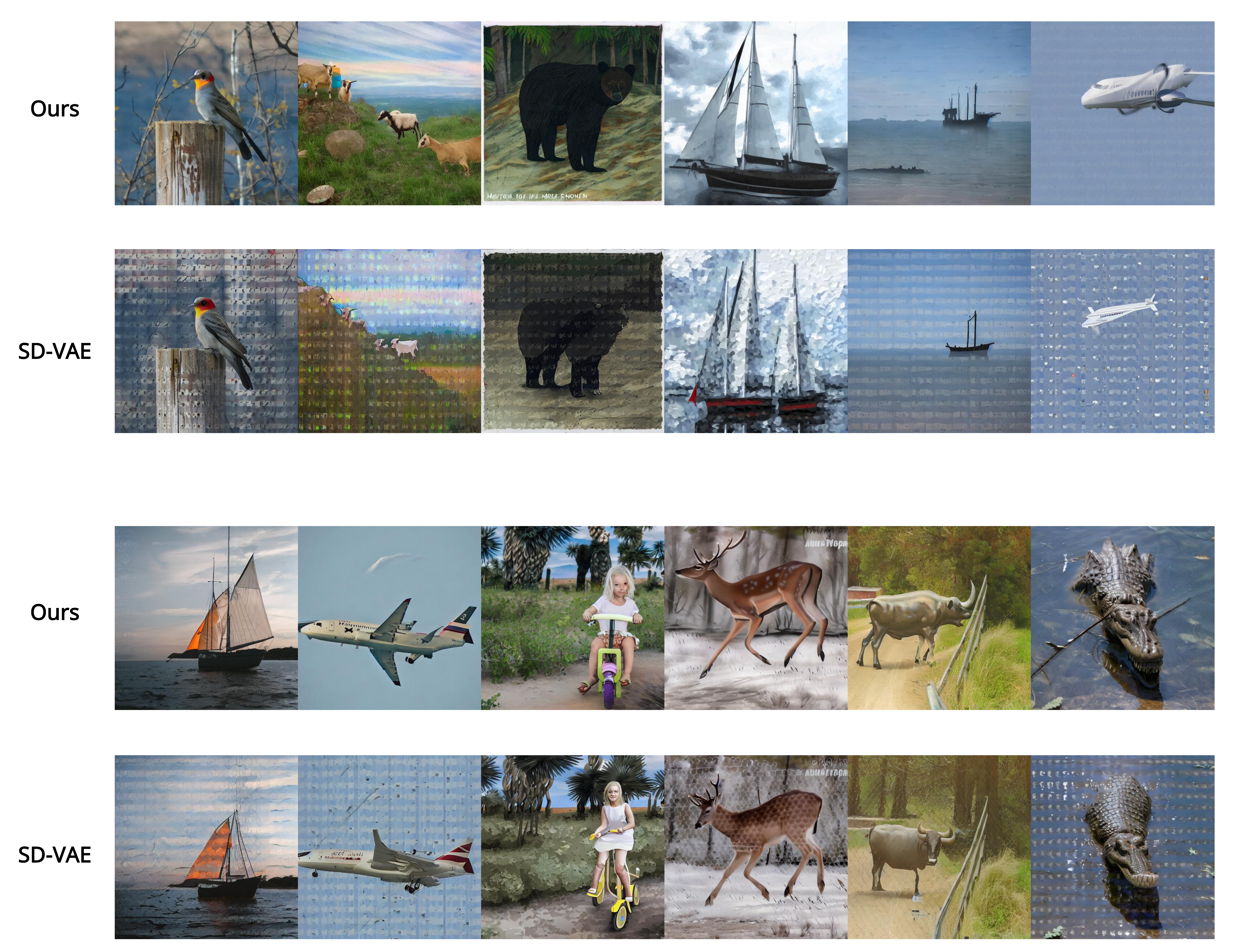}
    \vspace{-0.20in}
    \caption{\textbf{Comparison of image-to-image editing results on protected images (MIST).}}
    \vspace{-0.05in}
    \label{figure:app_editing_mist}
\end{figure*}

\begin{figure*}[t]
    \centering
    \includegraphics[width=1.0\linewidth]{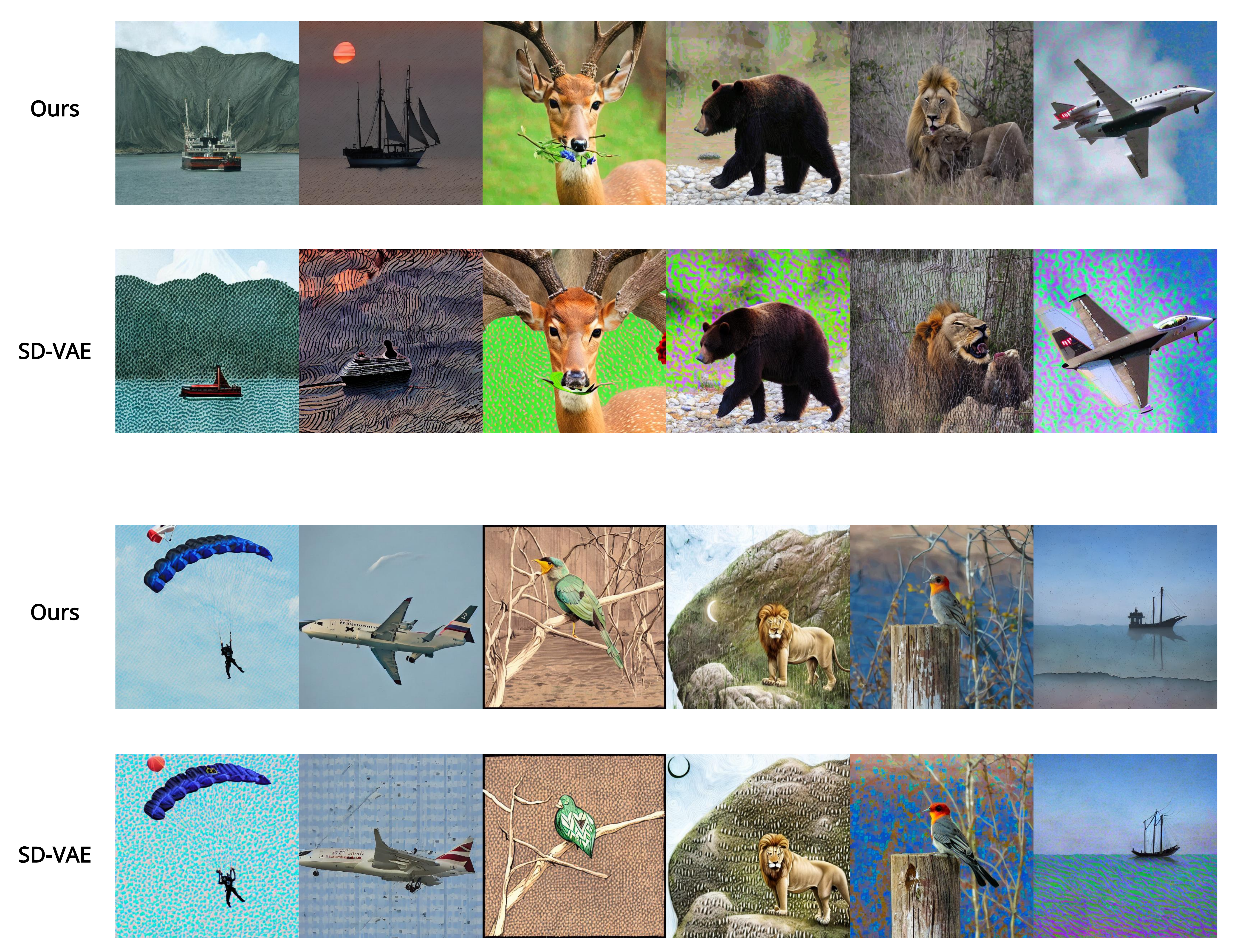}
    \vspace{-0.20in}
    \caption{\textbf{Comparison of image-to-image editing results on protected images (Glaze).}}
    \vspace{-0.05in}
    \label{figure:app_editing_glaze}
\end{figure*}

We provide additional qualitative results in our experiments. Figure~\ref{figure:app_nightshade} shows results under the Nightshade attack. While the baseline model produces manipulated outputs, SRL-VAE preserves the original prompt concept, demonstrating strong robustness. Figure~\ref{figure:app_editing_photoguard},~\ref{figure:app_editing_mist} and \ref{figure:app_editing_glaze} show image-to-image editing results on adversarially protected images, comparing the outputs of SRL-VAE and the baseline SD-VAE. These examples further demonstrate the robustness of SRL-VAE in reconstructing and editing perturbed inputs while preserving semantic consistency and visual quality.

\end{document}